\pdfoutput=1
\documentclass[11pt]{article}
\PassOptionsToPackage{table}{xcolor}
\PassOptionsToPackage{hyperfootnotes=false}{hyperref}
\usepackage[preprint]{acl}

\usepackage{times}
\usepackage{latexsym}
\usepackage[T1]{fontenc}
\usepackage[utf8]{inputenc}
\usepackage{microtype}
\usepackage{inconsolata}
\usepackage{graphicx}
\usepackage{booktabs}
\usepackage{amsmath}
\usepackage{amssymb}
\usepackage{amsthm}
\providecommand{\texorpdfstring}[2]{#1}
\DeclareRobustCommand{\benchmarkname}{\texorpdfstring{\textsc{Ekphrasis}}{Ekphrasis}}
\usepackage{array}
\usepackage{tabularx}
\usepackage{makecell}
\usepackage{adjustbox}
\usepackage{threeparttable}
\usepackage{siunitx}
\usepackage{placeins}
\usepackage{stfloats}
\usepackage{algorithm}
\usepackage{algpseudocode}
\usepackage[most]{tcolorbox}
\newtcolorbox{definitionbox}[1][]{
  colback=green!2, colframe=green!45!black,
  boxrule=0.4pt, arc=1pt, left=6pt, right=6pt, top=4pt, bottom=4pt,
  breakable, #1
}
\newtcolorbox{methodbox}[1][]{
  colback=blue!2, colframe=blue!45!black,
  boxrule=0.4pt, arc=1pt, left=6pt, right=6pt, top=4pt, bottom=4pt,
  breakable, #1
}
\newcolumntype{Y}{>{\raggedright\arraybackslash}X}
\newcolumntype{L}[1]{>{\raggedright\arraybackslash}p{#1}}
\newcolumntype{C}[1]{>{\centering\arraybackslash}p{#1}}
\newtheorem{definition}{Definition}
\definecolor{VCIGreen}{HTML}{0B7A53}
\definecolor{VCIRed}{HTML}{B23A48}
\definecolor{VCIGray}{HTML}{6B7280}
\definecolor{VCIBand}{HTML}{F1F3F5}
\definecolor{VCITableBand}{HTML}{E4E2DD}
\definecolor{VCITableAvg}{HTML}{F4F1E8}
\definecolor{VCITableOverall}{HTML}{EAF0F4}

\newcommand{\figureoneblock}{%
  \begin{figure*}[!t]
  \centering
  \includegraphics[width=\textwidth]{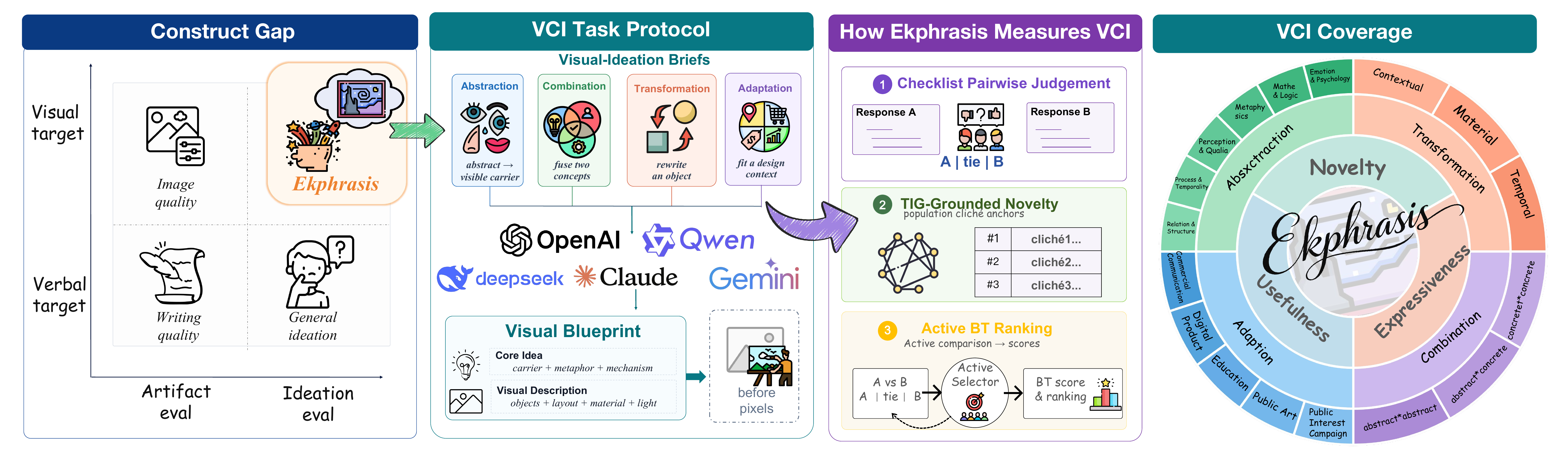}
  \caption{\textbf{Construct and evaluation workflow of \benchmarkname{}.} \benchmarkname{} isolates Visual Creative Ideation as a text-only, pre-image planning construct distinct from image quality, writing quality, and general ideation. The workflow elicits visual blueprints across four brief families, scores plans through checklist pairwise judgments with Typed Idea Graph-grounded novelty and active Bradley--Terry aggregation, and summarizes coverage across usefulness, expressiveness, novelty, and task subtypes.}
  \label{fig:benchmark-landscape}
  \end{figure*}
}

\newcommand{\figuretwoblock}{%
  \begin{figure*}[!t]
  \centering
  \includegraphics[width=\textwidth]{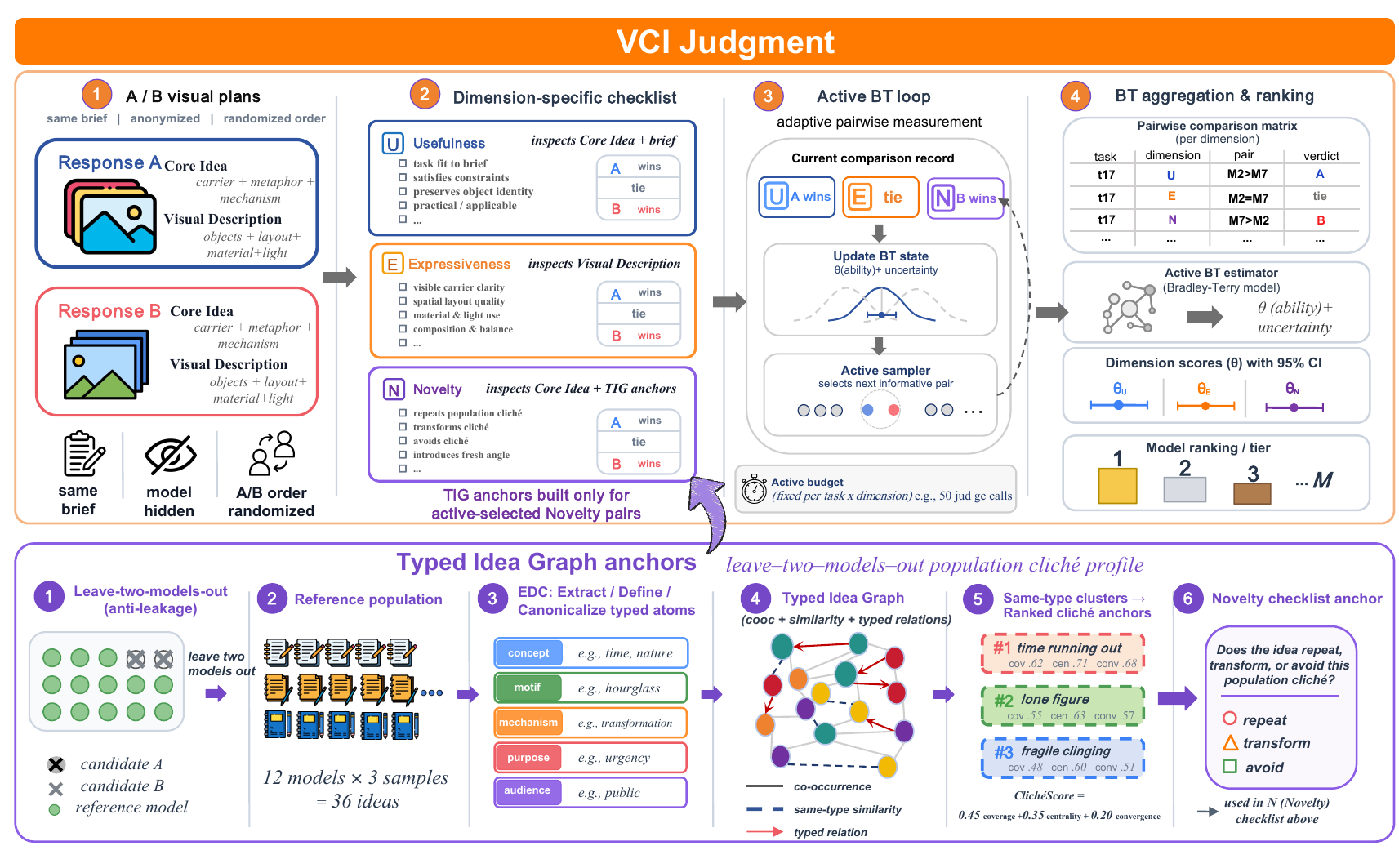}
  \caption{\textbf{Evaluation-method overview.} \benchmarkname{} anonymizes paired visual plans, evaluates them with dimension-specific checklists for Usefulness, Expressiveness, and Novelty, aggregates pairwise outcomes with Bradley--Terry models, and constructs population-anchored Novelty references through leave-two-models-out Typed Idea Graphs.}
  \label{fig:method-overview}
  \end{figure*}
}

\title{Can Language Models Imagine Without Seeing?\\
\benchmarkname{}: Measuring Visual Creative Ideation in Text-Only LLMs%
\thanks{\href{https://github.com/Imhongyu/Ekphrasis}{github.com/Imhongyu/Ekphrasis}}}

\author{
  \textbf{Hongyu Luo, He Wang, Huihao Jing, Hong Ting Tsang, Yuxuan Liu,} \\
  \textbf{Wuganjing Song, Yauwai Yim, Chunyang Li, Yangqiu Song} \\
  {\normalsize\normalfont The Hong Kong University of Science and Technology, Hong Kong SAR, China} \\
  {\footnotesize\normalfont
    \texttt{\{hluoay,hwangje,hjingaa,httsangaj,yliurk,wsongan,ywyimaa,cliej\}@connect.ust.hk},
    \texttt{yqsong@cse.ust.hk}}
}

\begin{document}
\maketitle

\begin{abstract}
Current evaluations do not isolate whether text-only language models
can originate visual concepts before image generation. Fluent visual
prose can hide visual-plan failures: an answer may appear creative
while repeating familiar visual clich\'es or failing to specify a
renderable scene. We define Visual Creative Ideation (VCI) as the
ability to produce textual visual plans that are useful, expressive, and
population-novel, and introduce \benchmarkname{}, a 400-task benchmark
spanning Abstraction, Combination, Transformation, and Adaptation.
\benchmarkname{} scores anonymized pairwise comparisons with
dimension-specific checklists, aggregates preferences with
Bradley--Terry models, and uses Typed Idea Graphs to convert
task-specific population clich\'es into novelty references. Across
14 language models, VCI separates usefulness, expressiveness, and
novelty rather than reducing to fluency: strong models achieve similar
overall scores through different profiles, and useful plans can remain
visually clich\'ed. A cross-modal grounding study further shows that
text-level VCI ordering largely survives faithful rendering and blind
image-level preference judgment, supporting \benchmarkname{} as a
measure of visual ideation beyond prose quality.
\end{abstract}

\figureoneblock

\section{Introduction}

Before an image is rendered, many creative decisions have already been
made. In concept-art brainstorming, advertising, pre-visualization, and
text-to-image workflows, a text-only LLM may choose the scene, carrier,
composition, and design direction that downstream artists, users, or
renderers later realize. The product is therefore neither an image nor
ordinary prose, but a \textit{pre-image visual plan}: language that must
carry visual substance.

This setting creates a measurement trap. A model can write vivid visual
language while recycling safe motifs, familiar composition recipes, and
recurring LLM-population clich\'es. We call this
\textit{novelty illusion}: outputs read as creative even when their
visual invention is thin. The problem is hard to detect in text alone
because prose can explain away what an image could not sustain.

Consider a brief about time pressure. A fluent plan might use an
hourglass, storm clouds, and a figure running out of sand. It fits the
brief but repeats a familiar carrier. Another plan might turn a subway
map into branching deadlines, with stations collapsing into missed
opportunities. The difference is not polish but mechanism: the latter
changes how time pressure becomes visible.

Existing evaluations largely miss this construct. Adjacent
benchmarks---verbal
creativity~\citep{guilford1967nature,olson2021naming,chakrabarty2024artifice,fein2025litbench,paech2025eqbenchcreative},
image-generation and image-editing artifacts~\citep{huang2023t2icompbench,ghosh2023geneval,wu2023hpsv2,zhao2025risebench,wu2025krisbench,han2025unireditbench},
visual reasoning~\citep{antol2015vqa,hudson2019gqa}, and instruction
following~\citep{zhou2023instructionfollowing}---target final texts,
rendered images, or constraint adherence
(Figure~\ref{fig:benchmark-landscape}), but do not isolate whether a
text-only plan is useful, imageable, and non-clich\'ed before pixels
exist.

\begin{center}
\textit{Can language models imagine visually, or only write as if they can?}
\end{center}

We introduce \textit{Visual Creative Ideation} (VCI): the ability to
originate textual visual plans that are \textit{useful} for a brief,
\textit{expressive} enough to support a concrete image, and
\textit{novel} relative to task-specific patterns in the LLM
population. This construction follows the standard view that creativity
combines originality with usefulness or effectiveness~\citep{runco2012standard}.
The object of evaluation is an externalized plan, not mental imagery,
prompt-engineering syntax, or final image quality. Cross-modal validation
then asks whether the text-level signal survives when the idea is
rendered and judged without the original prose.

Our contributions are:

\begin{itemize}\itemsep0pt
\item \textbf{(Construct)} We define VCI as pre-image visual planning
that separates brief satisfaction, visual specificity, and
population-novel departure from recurring LLM clich\'es.

\item \textbf{(Measurement)} We release \benchmarkname{}, a
400-instance suite spanning \textit{Abstraction}, \textit{Combination},
\textit{Transformation}, and \textit{Adaptation}, with
checklist-assisted Bradley--Terry comparison~\citep{bradley1952rank}
and Typed Idea Graphs for population-anchored Novelty.

\item \textbf{(Evidence and validation)} We evaluate 14 language models,
show that VCI is multidimensional rather than a fluency leaderboard, and
validate a faithful rendered subset with blind image-level human
preferences.
\end{itemize}

\figuretwoblock

\section{Related Work}

\benchmarkname{} occupies the visual-ideation cell in
Table~\ref{tab:benchmark-positioning}: the output is text, but the
target construct is visual concept origination before image generation.
Prior benchmarks cover adjacent cells without isolating text-only
prompt-side visual ideation with population-anchored novelty and
cross-modal validity evidence.

\subsection{Creativity Evaluation in Language Models}

Language-model creativity benchmarks mostly evaluate verbal artifacts,
procedural invention, or associative distance. Divergent-thinking tasks
such as Alternative Uses and Divergent Association probe originality
\citep{guilford1967nature,olson2021naming}, while recent benchmarks
extend to physical problem solving, code creativity, marketing ideation,
and creative writing
\citep{tian2024macgyver,lu2024neogauge,hou2025creativityprism,bhat2025creativitybenchmark,chakrabarty2024artifice,fein2025litbench,paech2025eqbenchcreative}.
These tasks motivate preference-based evaluation, but a strong answer
need not specify a renderable scene. \benchmarkname{} instead evaluates
an externalized plan for an absent image, and defines novelty relative
to recurring visual patterns in the model population for the same brief.

\subsection{Visual and Multimodal Evaluation}

Visual benchmarks usually score artifacts or understanding. Text-to-image
benchmarks such as T2I-CompBench, GenEval, and HPS v2 evaluate rendered
images for compositionality, alignment, or human preference
\citep{huang2023t2icompbench,ghosh2023geneval,wu2023hpsv2}; related
image--text compatibility and faithfulness metrics evaluate whether an
image preserves its textual input~\citep{hessel2021clipscore,hu2023tifa}.
Visual reasoning and instruction-following benchmarks test image
understanding or constraint adherence
\citep{antol2015vqa,hudson2019gqa,zhou2023instructionfollowing}. Recent
multimodal creativity benchmarks study creative intelligence with image
inputs or visual assets \citep{fang2025creationmmbench,xia2025visiar}.
Image-editing benchmarks such as SmartEdit, RISEBench, KRISBench, and
UniREditBench add stronger reasoning coverage and, in UniREditBench,
dual image/text references
\citep{huang2024smartedit,zhao2025risebench,wu2025krisbench,han2025unireditbench},
but still score edited images. \benchmarkname{} evaluates the plan that
precedes any image artifact.

\subsection{Pairwise Evaluation and LLM-as-Judge}

Open-ended creative outputs rarely admit a single gold answer, making
absolute ratings sensitive to preference and surface fluency. Pairwise
comparison with Bradley--Terry aggregation provides local judgments and
model-level scores \citep{bradley1952rank}, including in creative-writing
and marketing-ideation evaluation
\citep{fein2025litbench,bhat2025creativitybenchmark}. Because
LLM-as-judge systems can show position, verbosity, and fluency biases
\citep{zheng2023judging}, \benchmarkname{} uses construct-derived
checklists, reports the three VCI dimensions separately, anchors novelty
to a leave-out task population, and calibrates scalable judgments
against human annotations.

\begin{table*}[!t]
\centering
\caption{Coverage matrix positioning \benchmarkname{} relative to adjacent benchmark families. \benchmarkname{} is designed for text-only, pre-image visual plans, with population-anchored novelty and a separate image-level validation path.}
\label{tab:benchmark-positioning}
\scriptsize
\setlength{\tabcolsep}{3.2pt}
\renewcommand{\arraystretch}{1.04}
\begin{threeparttable}
\begin{adjustbox}{max width=\textwidth}
\begin{tabularx}{\textwidth}{@{}L{0.22\textwidth}L{0.16\textwidth}ccccccc@{}}
\toprule
\textbf{Benchmark family} & \textbf{Examples} &
\multicolumn{3}{c}{\textbf{Subject / output}} &
\multicolumn{3}{c}{\textbf{Construct}} &
\textbf{Validation} \\
\cmidrule(lr){3-5}\cmidrule(lr){6-8}\cmidrule(l){9-9}
& &
\makecell{\textbf{Text-only}\\\textbf{input}} &
\makecell{\textbf{Text plan}\\\textbf{scored}} &
\makecell{\textbf{Image}\\\textbf{artifact scored}} &
\makecell{\textbf{Visual}\\\textbf{target}} &
\makecell{\textbf{Pre-image}\\\textbf{ideation}} &
\makecell{\textbf{Population}\\\textbf{novelty}} &
\makecell{\textbf{Image-level}\\\textbf{check}} \\
\midrule
Verbal creativity / writing
& AUT / DAT; LitBench
& \checkmark & -- & -- & -- & -- & -- & -- \\
Text-to-image generation
& T2I-CompBench; GenEval
& \checkmark & -- & \checkmark & \checkmark & -- & -- & -- \\
Visual reasoning
& VQA; GQA
& -- & -- & -- & \checkmark & -- & -- & -- \\
Multimodal creativity
& Creation-MMBench; VISIAR
& -- & -- & \checkmark & \checkmark & -- & -- & -- \\
Image editing
& UniREditBench; KRISBench
& -- & -- & \checkmark & \checkmark & -- & -- & \checkmark \\
\midrule
\rowcolor{VCIBand}
\textbf{\benchmarkname{}}
& \textbf{Text-only visual briefs}
& \checkmark & \checkmark & -- & \checkmark & \checkmark & \checkmark & \checkmark \\
\bottomrule
\end{tabularx}
\end{adjustbox}
\begin{tablenotes}[flushleft]\footnotesize
\item Examples are representative; the surrounding related-work text gives the full cited set. A checkmark denotes a primary evaluation target, except that \benchmarkname{} uses image-level checking only for cross-modal validation rather than official text-level scoring.
\end{tablenotes}
\end{threeparttable}
\end{table*}

\section{Visual Creative Ideation and \benchmarkname{}}
\label{sec:ekphrasis}

\benchmarkname{} evaluates prompt-side visual ideation before pixels
are generated. Inputs and outputs remain textual, but the score targets
whether the plan fits the brief, supports a concrete image, and avoids
common LLM-population solutions.

\subsection{Task and Response Format}
\label{sec:ekphrasis_prelim}

\begin{definition}[Visual Creative Ideation]
\label{def:vci}
Given a visual brief $b$, \emph{Visual Creative Ideation} (VCI) is the
ability to originate a textual visual plan that is useful for the brief,
expressive enough to support a stable image, and non-clich\'ed with
respect to common responses to the same brief.
\end{definition}

An \benchmarkname{} task instance is a visual brief
\[
b=(x,\tau,C), \qquad b\in\mathcal{B},
\]
where $x$ is the input content, $\tau$ is the visual-ideation operation,
and $C$ denotes contextual or medium constraints. Given $b$, model $m$
returns
\[
p_m(b)=\big(i_m(b),d_m(b)\big),
\]
where the \emph{Core Idea} $i_m(b)$ states the carrier, metaphor or
mechanism, and key state or action, and the \emph{Visual Description}
$d_m(b)$ states what would be visible in one bounded static frame.

\subsection{Protocol and Evaluation Dimensions}
\label{sec:ekphrasis_protocol}

\begin{definition}[\benchmarkname{} Task Protocol]
\label{def:ekphrasis_protocol}
Given a brief $b\in\mathcal{B}$ and a language model $m$, the
\benchmarkname{} protocol requires $m$ to output a two-field textual
visual plan
\[
p_m(b)=\big(i_m(b),d_m(b)\big),
\]
where $i_m(b)$ states the core visual idea and $d_m(b)$ provides a
concrete visual description of the same idea.
\end{definition}

A valid response describes one static image and excludes tool names,
renderer parameters, multi-image narratives, and prompt-engineering
syntax. The two-field format prevents fluent scene prose from replacing
an appropriate, imageable, and non-obvious visual plan.

\begin{definition}[VCI Evaluation Dimensions]
\label{def:vci_dimensions}
For a model response $p_m(b)$, \benchmarkname{} evaluates VCI along
\[
\mathrm{Eval}(m,b)
=
\big(
U_m(b), E_m(b), N_m(b)
\big),
\]
where $U_m(b)$ denotes \emph{Usefulness}, $E_m(b)$ denotes
\emph{Expressiveness}, and $N_m(b)$ denotes \emph{Novelty}.
Usefulness measures whether the idea satisfies the brief. Expressiveness
measures whether the description supports a concrete mental image.
Novelty measures whether the core idea departs from population-level
visual clich\'es for the same brief.
\end{definition}

Usefulness and Novelty primarily inspect $i_m(b)$; Expressiveness
primarily inspects $d_m(b)$. Thus VCI is not image quality, verbal
creativity, instruction following alone, or renderer-specific prompting.

\subsection{Task Design}
\label{sec:task_design}

\benchmarkname{} organizes its briefs into four controlled task families:
\[
\mathcal{B}
=
\mathcal{B}_{\mathrm{abs}}
\cup
\mathcal{B}_{\mathrm{comb}}
\cup
\mathcal{B}_{\mathrm{trans}}
\cup
\mathcal{B}_{\mathrm{adapt}},
\]
where the first three families specify the creative operation and
Adaptation specifies a realistic context while leaving the visual
strategy open. Table~\ref{tab:task_families} summarizes the sampling
control, balancing axis, and diagnostic failure mode for each family;
exact family and subtype counts are reported in Appendix~\ref{app:dataset_details}.
Source pools draw on concreteness norms~\citep{brysbaert2014concreteness},
SUBTLEX-US frequency estimates~\citep{brysbaert2009subtlexus},
THINGS object concepts~\citep{hebart2019things}, and Places365 scene
categories~\citep{zhou2018places365}.

\begin{table*}[t]
\centering
\caption{Task-family construction controls.}
\label{tab:task_families}
\footnotesize
\setlength{\tabcolsep}{4pt}
\renewcommand{\arraystretch}{1.1}
\begin{tabularx}{\textwidth}{@{}L{0.14\textwidth}Y L{0.27\textwidth}@{}}
\toprule
\textbf{Family} & \textbf{Construction control} & \textbf{Stress tested} \\
\midrule
Abstraction
& Curated abstract nouns; semantic subtype, abstractness, frequency, and visualizability controls
& Vague metaphor; decorative symbol \\
Combination
& Concrete--concrete, abstract--concrete, and abstract--abstract concept pairs; domain balance, distance band, and repetition cap
& Juxtaposition; trivial or incoherent fusion \\
Transformation
& Object pool crossed with material, temporal, and contextual specifications; category and distance balance
& Surface edit; loss of object identity \\
Adaptation
& Web-collected real-world visual-design needs rewritten as applied briefs; domain, medium, audience, and deliverable balance
& Generic campaign trope; missed constraint \\
\bottomrule
\end{tabularx}
\end{table*}

Abstraction tests abstract-to-visual grounding; Combination tests whether
models fuse concepts rather than merely juxtapose them; Transformation
tests condition-guided object rewriting under recognizability
constraints; and Adaptation tests visual strategy selection under
realistic communication or design constraints. Representative prompts
and subtype distributions are reported in Appendix~\ref{app:dataset_details}.

\subsection{Dataset Construction and Quality Control}
\label{sec:dataset_construction}

Tasks are constructed through template-based authoring, controlled
sampling, real-brief rewriting, and human curation. Each prompt is
screened for ambiguity, single-image renderability, family/subtype
overlap, sensitive or copyrighted content, and distributional balance.
A pilot pass then verifies that prompts elicit the required Core Idea and
Visual Description fields rather than essays, tool commands, multi-image
sequences, or non-visual explanations. Detailed sampling protocols and
prompt templates are given in Appendix~\ref{app:dataset_details}.

\section{Evaluation Method}
\label{sec:evaluation}

\benchmarkname{} scores open-ended visual plans through
checklist-assisted pairwise comparison. Usefulness and Expressiveness
use dimension-specific rubrics; Novelty uses task-specific clich\'e
references from leave-out Typed Idea Graphs. Pairwise preferences are
aggregated with Bradley--Terry (BT) models \citep{bradley1952rank}.
Figure~\ref{fig:method-overview} summarizes this measurement engine.

\subsection{Checklist-Assisted Pairwise Evaluation}
\label{sec:pairwise_eval}

For each brief $b$, evaluation dimension $d\in\{U,E,N\}$, and model
pair $(m,n)$, the judge receives anonymized outputs $p_m(b)$ and
$p_n(b)$ together with a dimension-specific checklist
$\mathcal{C}_{b,d}$, then returns
\[
J_d\!\left(b,p_m(b),p_n(b),\mathcal{C}_{b,d}\right)
\rightarrow
y^{m,n}_{b,d}\in\left\{1,\frac{1}{2},0\right\},
\]
where $1$ means that $m$ is preferred to $n$, $0$ means that $n$ is
preferred to $m$, and $\frac{1}{2}$ denotes a tie. Model identities are
hidden, A/B order is randomized, and swapped-order repeats estimate
position bias.

For each dimension, the BT model assigns model $m$ a latent skill
$\theta_{m,d}$ and models the probability that $m$ is preferred to $n$
as
\[
\Pr(m\succ n\mid d)
=
\sigma(\theta_{m,d}-\theta_{n,d}).
\]
Only decisive comparisons enter the main BT likelihood; ties are kept
for diagnostics. We fit one BT model per dimension and report
$\hat{\theta}_{m,U}$, $\hat{\theta}_{m,E}$, and $\hat{\theta}_{m,N}$.
Regularization, active sampling, and confidence intervals are specified
in Appendix~\ref{app:evaluation_details}.

\subsection{Typed Idea Graphs for Population-Anchored Novelty}
\label{sec:typed_idea_graph}

Novelty is task-relative: a response is novel when it avoids, recombines,
or transforms visual patterns that other models repeatedly use for the
same brief. For each judged pair $q=\{m,n\}$, we build a
leave-two-models-out reference population from all other models'
responses to that brief, preventing either candidate from defining the
clich\'e profile used to judge it.

Responses in the reference population are mapped into typed idea atoms
(\emph{concept}, \emph{motif}, \emph{mechanism}, \emph{purpose}, and
\emph{audience}), connected by co-occurrence, semantic similarity, and
typed relations, and clustered into population-common patterns.
The Novelty checklist is constructed as
\[
\mathcal{R}^{-q}_{b}
\xrightarrow{\;\mathrm{EDC}\;}
A^{-q}_{b}
\xrightarrow{\;\mathrm{graph}\;}
G^{-q}_{b}
\xrightarrow{\;\mathrm{cluster+score}\;}
\mathcal{C}^{q}_{b,N},
\]
where EDC denotes Extract--Define--Canonicalize. Each cluster is scored
by prevalence, structural centrality, and semantic convergence:
\[
S(c)
=
\alpha \bar{P}(c)
+
\beta \bar{C}(c)
+
\gamma \bar{Q}(c),
\qquad
\alpha+\beta+\gamma=1.
\]
High-scoring clusters are verbalized as checklist anchors. During
judging, the evaluator marks whether each candidate \textsc{repeats},
\textsc{transforms}, or \textsc{avoids} those anchors, so Novelty is
measured against task-specific visual patterns rather than generic
semantic distance. Extraction prompts, graph construction, and audit
criteria are detailed in Appendix~\ref{app:tig_details}.

\subsection{Score Reporting and Calibration}
\label{sec:score_reporting}

\begin{figure*}[!t]
\centering
\includegraphics[width=\textwidth]{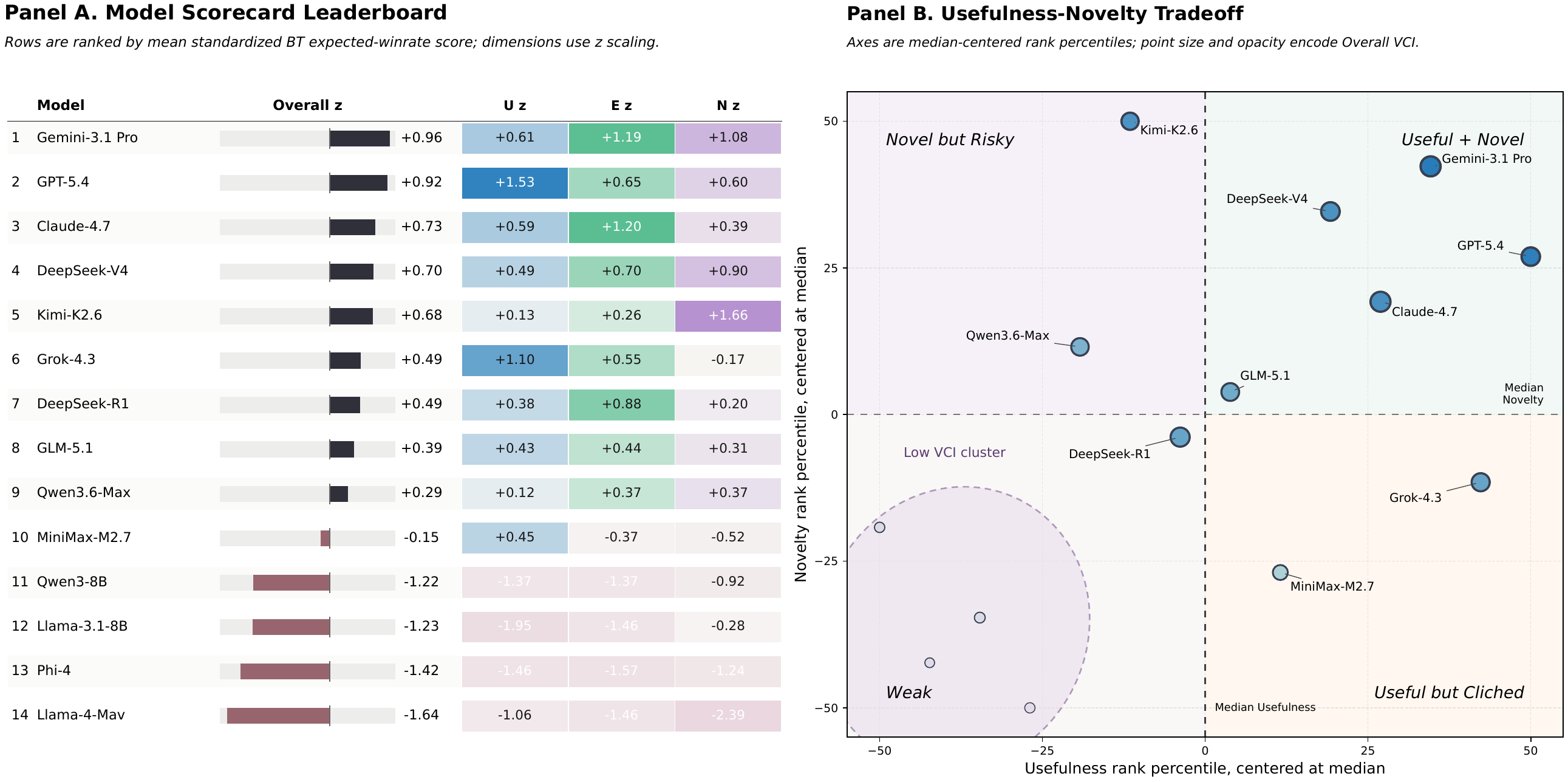}
\caption{Main VCI results. Panel A ranks models by Overall VCI, computed
as the mean of standardized BT expected-winrate scores across
Usefulness, Expressiveness, and Novelty. Bars show Overall VCI and
heatmap cells show dimension-specific $z$ scores. Panel B keeps the
Usefulness--Novelty tradeoff display using median-centered rank
percentiles for visualization only; point size and opacity encode
Overall VCI.}
\label{fig:main-results}
\end{figure*}

The main results use dimension-specific BT expected-winrate scores.
Because the three dimensions have different empirical spreads, we
standardize each dimension before computing Overall VCI:
\[
z_{m,d}=\frac{s_{m,d}-\mu_d}{\sigma_d},
\]
where $s_{m,d}$ is the BT expected-winrate score and $\mu_d,\sigma_d$
are the mean and standard deviation over subject models. Overall VCI is
the equal-weight mean of the three $z$ scores. For visualization only,
Figure~\ref{fig:main-results} uses median-centered rank percentiles to
show the Usefulness--Novelty tradeoff; these coordinates are not
absolute quality percentages.

We use a two-tier evaluation design: a scalable checklist-assisted BT
judge produces the full leaderboard, while human annotations provide
targeted calibration and validation slices for agreement, reliability,
and cross-modal grounding.

\section{Experiments}
\label{sec:experiments}
We evaluate 14 text-only LLMs on the 400-task \benchmarkname{} suite.
The official leaderboard uses the first pre-registered sample for each
model--task pair; additional samples only estimate task-specific
population clich\'es for Typed Idea Graph construction.

\subsection{Experimental Setup}
Usefulness, Expressiveness, and Novelty are judged separately with
checklist-assisted pairwise comparison and one BT model per dimension.
The official DeepSeek-V4-Flash scorer uses active BT sampling: after
warm-start coverage, it prioritizes comparisons that are informative
under current uncertainty. Order swaps, duplicate checks, auxiliary
judges, and human validation slices characterize reliability rather
than redefine the leaderboard. Budgets, pilot criteria, and validation
scale are reported in Appendix~\ref{app:experimental_matrix}.

\subsection{Main Results}

Figure~\ref{fig:main-results} summarizes the primary evidence. Gemini
3.1 Pro (+0.96), GPT-5.4 (+0.92), Claude 4.7 (+0.73), DeepSeek V4
(+0.70), and Kimi K2.6 (+0.68) form a close, profile-diverse top tier.
GPT-5.4 leads Usefulness (+1.53), Claude 4.7 and Gemini 3.1
Pro lead Expressiveness (+1.20/+1.19), and Kimi K2.6 leads Novelty
(+1.66). Grok 4.3 shows the dissociation: high Usefulness (+1.10) can
coexist with below-average Novelty (-0.17).

\subsection{Model Profiles and Task-Family Stress Tests}

Task families reveal profile-specific strengths: Gemini 3.1 Pro
leads Abstraction (+1.03), GPT-5.4 leads Adaptation (+1.04), and Kimi
K2.6 leads Transformation (+1.07). Combination is most demanding
because it requires shared visual structure rather than side-by-side
placement, making it the clearest family-level stress test.
Table~\ref{tab:model-task-family-performance} reports the full
model-by-family breakdown.

\begin{table*}[!t]
\centering
\caption{Model performance by task family. Cells are family-specific standardized Bradley--Terry scores; higher is better, and bold marks the row-best model.}
\label{tab:model-task-family-performance}
\scriptsize
\setlength{\tabcolsep}{3.0pt}
\renewcommand{\arraystretch}{1.08}
\begin{threeparttable}
\begin{adjustbox}{max width=\textwidth}
\begin{tabular}{@{}l*{14}{S[table-format=+1.2]}@{}}
\toprule
\textbf{Metric} & \multicolumn{1}{c}{\textbf{Gemini}} & \multicolumn{1}{c}{\textbf{GPT-5.4}} & \multicolumn{1}{c}{\textbf{Claude}} & \multicolumn{1}{c}{\textbf{DeepSeek}} & \multicolumn{1}{c}{\textbf{Kimi}} & \multicolumn{1}{c}{\textbf{Grok}} & \multicolumn{1}{c}{\textbf{R1}} & \multicolumn{1}{c}{\textbf{GLM-5.1}} & \multicolumn{1}{c}{\textbf{Qwen-Max}} & \multicolumn{1}{c}{\textbf{MiniMax}} & \multicolumn{1}{c}{\textbf{Qwen-8B}} & \multicolumn{1}{c}{\textbf{Llama-8B}} & \multicolumn{1}{c}{\textbf{Phi-4}} & \multicolumn{1}{c}{\textbf{Llama-Mav}} \\
\midrule
\rowcolor{VCITableBand}\multicolumn{15}{@{}l}{\textbf{Abstraction}\hspace{0.45em}\textit{abstraction}} \\
U & +0.85 & \bfseries +1.42 & +0.75 & +0.58 & -0.06 & +0.77 & -0.13 & +0.59 & -0.00 & +0.85 & -1.31 & -2.13 & -1.27 & -0.91 \\
E & +0.99 & +0.65 & \bfseries +1.05 & +0.63 & +0.34 & +0.50 & +0.85 & +0.51 & +0.61 & -0.03 & -1.57 & -1.53 & -1.60 & -1.42 \\
N & \bfseries +1.26 & +0.93 & +0.54 & +0.64 & +1.25 & -0.27 & -0.01 & +0.10 & +0.57 & -0.63 & -0.47 & -0.17 & -1.08 & -2.66 \\
\rowcolor{VCITableAvg}\textit{Task avg.} & \bfseries +1.03 & +1.00 & +0.78 & +0.62 & +0.51 & +0.34 & +0.24 & +0.40 & +0.39 & +0.07 & -1.12 & -1.27 & -1.32 & -1.66 \\
\rowcolor{VCITableBand}\multicolumn{15}{@{}l}{\textbf{Combination}\hspace{0.45em}\textit{combination}} \\
U & +0.67 & \bfseries +1.22 & +0.01 & +0.68 & +0.11 & +1.06 & +0.61 & +0.71 & +0.37 & +0.43 & -1.08 & -2.03 & -1.67 & -1.10 \\
E & \bfseries +1.12 & +0.92 & +0.83 & +0.69 & +0.23 & +0.59 & +0.76 & +0.65 & +0.47 & -0.27 & -1.24 & -1.59 & -1.63 & -1.53 \\
N & +0.12 & +0.76 & +0.57 & +0.78 & \bfseries +1.77 & +0.19 & +0.15 & +0.70 & +0.30 & -0.34 & -1.10 & -0.28 & -1.22 & -2.41 \\
\rowcolor{VCITableAvg}\textit{Task avg.} & +0.63 & \bfseries +0.97 & +0.47 & +0.72 & +0.70 & +0.62 & +0.51 & +0.69 & +0.38 & -0.06 & -1.14 & -1.30 & -1.50 & -1.68 \\
\rowcolor{VCITableBand}\multicolumn{15}{@{}l}{\textbf{Transformation}\hspace{0.45em}\textit{transformation}} \\
U & +0.54 & +0.96 & +0.87 & +0.62 & +0.58 & \bfseries +1.08 & +0.67 & +0.40 & +0.01 & +0.26 & -1.70 & -1.78 & -1.62 & -0.88 \\
E & +0.90 & +0.76 & \bfseries +1.14 & +0.72 & +0.42 & +0.52 & +0.75 & +0.54 & +0.41 & -0.05 & -1.36 & -1.56 & -1.65 & -1.52 \\
N & +0.94 & -0.49 & +0.10 & +0.47 & \bfseries +2.20 & -0.27 & +0.59 & -0.08 & +0.60 & -0.02 & -0.85 & -0.04 & -0.78 & -2.38 \\
\rowcolor{VCITableAvg}\textit{Task avg.} & +0.80 & +0.41 & +0.70 & +0.60 & \bfseries +1.07 & +0.44 & +0.67 & +0.29 & +0.34 & +0.06 & -1.30 & -1.13 & -1.35 & -1.59 \\
\rowcolor{VCITableBand}\multicolumn{15}{@{}l}{\textbf{Adaptation}\hspace{0.45em}\textit{adaptation}} \\
U & +0.20 & \bfseries +2.12 & +0.54 & +0.12 & +0.04 & +1.26 & +0.44 & -0.04 & +0.27 & +0.18 & -1.14 & -1.99 & -1.03 & -0.97 \\
E & \bfseries +0.91 & +0.36 & +0.89 & +0.73 & +0.63 & +0.80 & +0.83 & +0.48 & +0.48 & +0.01 & -1.38 & -1.39 & -1.81 & -1.55 \\
N & +1.27 & +0.63 & +0.25 & +1.27 & \bfseries +1.51 & -0.09 & +0.14 & +0.52 & -0.05 & -0.64 & -0.84 & -0.48 & -1.38 & -2.12 \\
\rowcolor{VCITableAvg}\textit{Task avg.} & +0.80 & \bfseries +1.04 & +0.56 & +0.71 & +0.73 & +0.66 & +0.47 & +0.32 & +0.23 & -0.15 & -1.12 & -1.29 & -1.41 & -1.55 \\
\midrule
\rowcolor{VCITableOverall}\textit{Overall avg.} & +0.82 & \bfseries +0.85 & +0.63 & +0.66 & +0.75 & +0.51 & +0.47 & +0.42 & +0.34 & -0.02 & -1.17 & -1.25 & -1.39 & -1.62 \\
\bottomrule
\end{tabular}
\end{adjustbox}
\begin{tablenotes}[flushleft]\footnotesize
\item U, E, and N denote Usefulness, Expressiveness, and Novelty. Model abbreviations: Gemini = Gemini 3.1 Pro; Claude = Claude Opus 4.7; DeepSeek = DeepSeek V4 Pro; R1 = DeepSeek R1; Qwen-Max = Qwen 3.6-Max; MiniMax = MiniMax M2.7; Llama-8B = Llama 3.1 8B Instruct; Llama-Mav = Llama 4 Maverick.
\end{tablenotes}
\end{threeparttable}
\end{table*}%

\subsection{Novelty Ablation}

To test whether task-specific Typed Idea Graph (TIG) anchors are needed
for Novelty judgment, we rerun the same human gold subset with generic
anchors in place of case-specific checklist items. TIG anchoring
improves human-majority agreement by 20.9 percentage points
(Table~\ref{tab:novelty-ablation}).

\begin{table}[!t]
\centering
\caption{\textbf{TIG anchors improve novelty agreement.} Human-aligned agreement is measured on 235 paired Novelty comparisons.}
\label{tab:novelty-ablation}
\footnotesize
\setlength{\tabcolsep}{7pt}
\renewcommand{\arraystretch}{1.08}
\begin{tabular}{@{}lcc@{}}
\toprule
\textbf{Condition} & \textbf{Correct / $n$} & \textbf{Agreement} \\
\midrule
TIG anchors & 180 / 235 & 76.6\% \\
No TIG & 131 / 235 & 55.7\% \\
\bottomrule
\end{tabular}
\par\smallskip
\begin{minipage}{0.94\columnwidth}
\footnotesize\raggedright
Gain, \textbf{+20.9 pp} (95\% CI, [+14.0, +28.1] pp); discordant correct, 64 versus 15; McNemar $p<0.001$.
\end{minipage}
\end{table}

\subsection{Reliability Checks}
\label{sec:reliability}

Reliability diagnostics support the model ranking: JSON repair is rare,
tie rates are non-degenerate, order-swap rank correlations remain high,
and auxiliary judges recover broadly similar rankings. Length remains a
residual confound, especially for Expressiveness; dimension-wise
diagnostics are reported in Appendix~\ref{app:reliability_details}.

\section{Cross-Modal Grounding Validation}
We further test whether text-level VCI scores capture visual content
rather than only verbal quality. Textual plans are rendered with a
fixed text-to-image system, screened for faithfulness to the source
idea, and then compared by humans using only the rendered images. The
resulting image-level preferences are compared with the original
text-level VCI rankings, following prior emphasis on image--text
compatibility, text-to-image faithfulness, and reproducible human
evaluation for generated images~\citep{hessel2021clipscore,hu2023tifa,otani2023verifiable}.

\subsection{Faithfulness Gate}
The faithfulness gate ensures that image-level validation is performed
only on renderings that preserve the source visual plan. A rendering
passes when it preserves the central carrier, objects, relation, and
context needed for image-level comparison; hard failures include
semantic drift, concept splitting, object identity loss, and
style/context mismatch. In the completed annotation pass, each of the
240 rendered images receives three independent faithfulness annotations,
for 720 judgments in total. Majority vote passes 179 of 240 renderings
(74.6\%). The mean faithfulness score is 3.86/5; pass/fail agreement is
substantial (Fleiss $\kappa=0.637$), and ordinal score reliability is
high for a visual screening task (ICC $=0.711$), using standard
multi-rater reliability measures~\citep{fleiss1971measuring,shrout1979intraclass}.

\subsection{Image-Level Preference Agreement}
For the faithful subset, humans compare rendered images without seeing
the original text, model identity, or text-level ranking. The completed
image-preference subset contains 160 image pairs and 480 judgments.
Annotators reach a majority preference on 158 of 160 pairs (98.8\%),
with 73.8\% unanimous agreement and Fleiss $\kappa=0.690$. At the pair
level, decisive image preferences agree with the text-level model
ordering on 142 of 147 decisive pairs (96.6\%); counting image ties as
neutral gives 142 of 158 non-missing majority outcomes (89.9\%). We
therefore use the rendered-image study as convergent validity evidence,
not as a replacement for the text-level benchmark or as a separate
full-scale ranking benchmark.

\begin{figure*}[t]
\centering
\includegraphics[width=0.92\textwidth]{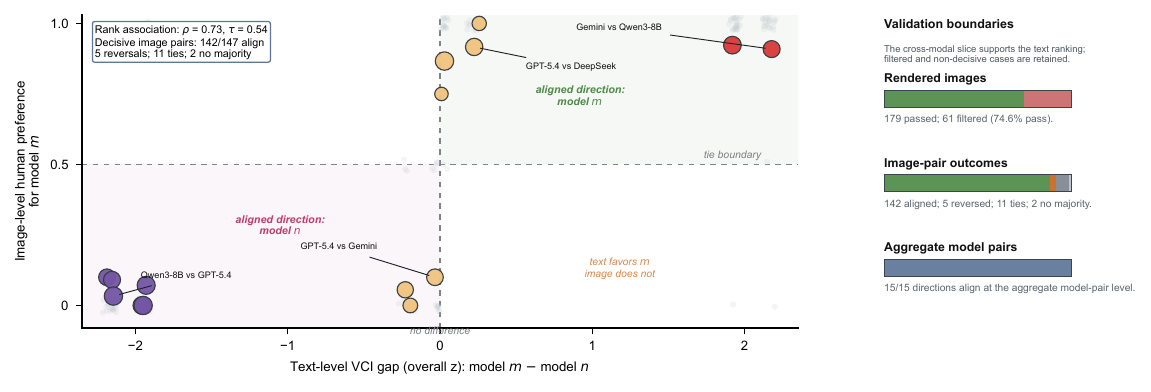}
\caption{Agreement between text-level VCI ordering and image-level human preferences in the six-model cross-modal subset. Faint points are individual image-pair majority outcomes; larger points summarize the corresponding model-pair preference rate. The horizontal axis shows the text-level Overall VCI gap, and the vertical axis shows the image-level human preference for model $m$.}
\label{fig:preference-agreement}
\end{figure*}

\section{Discussion}
\benchmarkname{} answers the title question with a qualified operational
yes: some text-only outputs specify visual plans whose structure remains
recoverable after rendering and blind image comparison. This is not
evidence of private mental imagery. It shows that pre-image plans can
carry task-relevant visual information across a text-to-image modality
shift. VCI is therefore not monolithic: similar
aggregate scores can reflect brief following, vivid specification, or
departure from population clich\'es, and plausible scenes can remain
visually familiar.

The central boundary exposed by the benchmark is safe plausibility.
Many outputs satisfy the brief and read well while falling back on
familiar carriers, stock metaphors, or predictable compositions. This
failure is subtler than missing a constraint because ordinary text
evaluation can reward it: the output appears thoughtful until it is
compared with the task-specific pattern of what other models also
produce. Combination tasks expose this most clearly when weak plans
place two concepts beside each other rather than inventing a shared
visual structure; Transformation tasks show a related surface edit in
which the object is decorated but its visible mode of existence is
unchanged.

Population-anchored Novelty is therefore not an ornamental creativity
score. Typed Idea Graphs make the reference set explicit: a plan is
judged against recurring solutions for the same brief, with the
candidate models left out of the clich\'e profile. The TIG ablation
suggests that this reference mainly improves human-aligned Novelty
judgments rather than acting as a generic preference boost. This matters
because novelty in VCI is relational: a plan can be semantically
appropriate and visually clear while still reproducing the model
population's default answer.

The cross-modal study provides a second boundary. It gives convergent
validity for the text-level VCI signal beyond verbal fluency, but only
for ideas that a fixed renderer preserves well enough to compare.
Rendering failures can reflect the image model as much as the source
plan; image-level agreement is validation evidence, not a
replacement image-generation leaderboard. More broadly, \benchmarkname{}
measures externalized pre-image planning, not private mental imagery or
a general theory of human creativity.

\section{Conclusion}
\benchmarkname{} operationalizes Visual Creative Ideation as useful,
expressive, and population-novel visual planning before pixels exist.
Across 400 tasks and 14 language models, it shows that brief
satisfaction, imageability, and escape from recurring visual clich\'es
are related but separable: strong systems reach similar overall VCI
through different profiles, and high usefulness can remain visually
familiar. The results support a scoped answer to the title question:
text-only models can specify imageable visual plans without direct image
input, but this ability must be tested against task fit, recoverability,
and the clich\'es of their own model population. For evaluation, the
practical implication is direct: creative-looking prose should not be
treated as visual creativity without separate checks for visual
mechanism, imageability, and population novelty.

\section*{Limitations}
\benchmarkname{} is limited by its task distribution, language and
cultural assumptions, and the current model population used to define
population-level clich\'es. Checklist-assisted LLM judgment reduces
annotation cost but still exhibits position and length effects, which
we report rather than eliminate. Cross-modal grounding also depends on
the chosen renderer: a failed rendering can reflect image-generation
limits rather than poor text-level ideation. Finally,
population-anchored Novelty is relative to the evaluated model pool and
should not be read as an absolute theory of human originality.

\section*{Ethical Considerations}
\benchmarkname{} scores should be used as diagnostic evidence about
model behavior, not as broad claims about human creativity or artistic
value. Novelty judgments may encode cultural assumptions about what
counts as a clich\'e, so released checklist materials and annotation
guidelines should be auditable. Human annotators should be informed
about task content, data use, and disagreement handling. Because the
benchmark can rank commercial systems, reporting should avoid
overclaiming beyond the tested task families and evaluation protocol.

\bibliography{custom}

\clearpage
\appendix

\section{Full Task Templates}
\label{app:dataset_details}
This appendix gives the operational task templates used to instantiate
\benchmarkname{}. The main paper describes the construct and task
families; here we make the elicitation format explicit enough for
replication.

\begin{definitionbox}
\textbf{Unified output format.} For every task $t$ and model $m$, the
expected response is a pair
\[
y_{m,t}=(c_{m,t}, v_{m,t}),
\]
where $c_{m,t}$ is a one-sentence \textit{Core Idea} and $v_{m,t}$ is a
100--140 word \textit{Visual Description}. The Core Idea states the
visual metaphor, carrier, state, or action; it should not spend its
budget on color, lighting, or atmosphere. The Visual Description states
only what is literally visible in a single static frame, binding visual
attributes such as color, light, texture, and mood to named elements.
\end{definitionbox}

Formally, a task instance is
\[
x_t=(f_t,b_t,C_t,S_t),
\]
where $f_t$ is the task family, $b_t$ is the natural-language brief,
$C_t$ is an optional set of constraints, and $S_t$ is the family-specific
source structure. The four task families differ in $S_t$ and in the
creative operation they elicit.

\begin{algorithm*}[!b]
\caption{Controlled task construction for \benchmarkname{}}
\label{alg:task-construction}
\begin{algorithmic}[1]
\Require Family set $\mathcal{F}$, source pools $\mathcal{P}_f$, target counts $n_f$
\Ensure Task set $\mathcal{T}$
\State $\mathcal{T}\gets \emptyset$
\ForAll{$f \in \mathcal{F}$}
  \State Construct subtype grid $\mathcal{G}_f$ and cell quotas
  \State Sample candidate variables $S_t\sim\mathcal{P}_f$ under each cell constraint
  \State Compose $x_t=(f,b_t,C_t,S_t)$
  \State Keep $x_t$ iff it is visual, unambiguous, safe, and non-duplicate
  \State Add validated instances until family quota $n_f$ is met
\EndFor
\State Assign stable ids and return balanced task set $\mathcal{T}$
\end{algorithmic}
\end{algorithm*}

\subsection{Abstraction}
Abstraction tasks elicit the operation
\[
\phi_{\mathrm{abs}}: a \mapsto v,
\]
where $a$ is an abstract target and $v$ is a concrete visual plan. The
prompt asks the model to express the abstract concept through a visible
object, scene, relation, or action rather than through text in the image.

\begin{methodbox}
\textbf{Template.}
\begin{quote}
\textbf{Task Type:} Abstraction\\
\textbf{Brief:} Design a visual concept that expresses the abstract
concept $a$.\\
\textbf{Output:} Give exactly two labeled blocks: Core Idea and Visual
Description.
\end{quote}
\end{methodbox}

The evaluation checks whether the response identifies a concrete carrier
for the target concept, whether the carrier maps specifically to the
brief rather than to a generic emotion or theme, and whether the final
image can be imagined without explanatory text.

\subsection{Combination}
Combination tasks elicit visual fusion:
\[
\phi_{\mathrm{comb}}:(c_1,c_2,\delta) \mapsto v,
\]
where $c_1$ and $c_2$ are source concepts and $\delta$ is a semantic
distance band. The intended output is a single integrated visual entity,
not a collage containing two unrelated source objects.

\begin{methodbox}
\textbf{Template.}
\begin{quote}
\textbf{Task Type:} Combination\\
\textbf{Brief:} Fuse $c_1$ and $c_2$ into one coherent visual entity.\\
\textbf{Constraint:} Both source concepts must jointly participate in
structure, function, or form.
\end{quote}
\end{methodbox}

The distance band controls how much conceptual bridging is required:
near pairs test subtle synthesis, medium pairs test nontrivial blending,
and far pairs test whether a model can invent a shared visual logic.

\subsection{Transformation}
Transformation tasks elicit object rewriting:
\[
\phi_{\mathrm{tfm}}:(o,r) \mapsto v,
\]
where $o$ is a source object and $r$ is a transformation condition. We
use three transformation subtypes: material, temporal, and contextual.

\begin{methodbox}
\textbf{Template.}
\begin{quote}
\textbf{Task Type:} Transformation\\
\textbf{Brief:} Re-present object $o$ under transformation condition
$r$.\\
\textbf{Constraint:} The condition should reshape the object itself
rather than merely decorate it or provide a background.
\end{quote}
\end{methodbox}

Material transformations require the new material to alter the object's
surface and structure; temporal transformations require visible state
change; contextual transformations require the environment's rules to
rewrite the object's mode of existence.

\subsection{Adaptation}
Adaptation tasks elicit applied visual problem solving:
\[
\phi_{\mathrm{adp}}:(s,C) \mapsto v,
\]
where $s$ is an applied situation and $C$ is a set of audience, medium,
domain, or communication constraints. In the implementation, the legacy
directory name is \texttt{adaption}; in the paper we use
\textit{Adaptation}.

\begin{methodbox}
\textbf{Template.}
\begin{quote}
\textbf{Task Type:} Adaptation\\
\textbf{Brief:} Produce a visual deliverable for applied situation $s$.\\
\textbf{Constraints:} Satisfy all explicit audience, medium, domain, and
communication constraints in $C$.
\end{quote}
\end{methodbox}

Unlike the first three families, Adaptation leaves the creative operation
less prescribed. This makes it useful for analyzing strategy choice, but
it also makes usefulness more sensitive to constraints.

\section{Dataset Statistics}
The released benchmark contains 400 task instances: 120 Abstraction,
90 Combination, 90 Transformation, and 100 Adaptation briefs.

\begin{definitionbox}
\textbf{Families are elicitation operations, not topics.} A task family
specifies the operation that the model must perform: abstract-to-visual
mapping, concept fusion, object rewriting, or constrained applied
design. The same topic can appear in different families, but it tests a
different form of VCI depending on the operation.
\end{definitionbox}

\begin{figure*}[!t]
\centering
\includegraphics[width=0.92\linewidth]{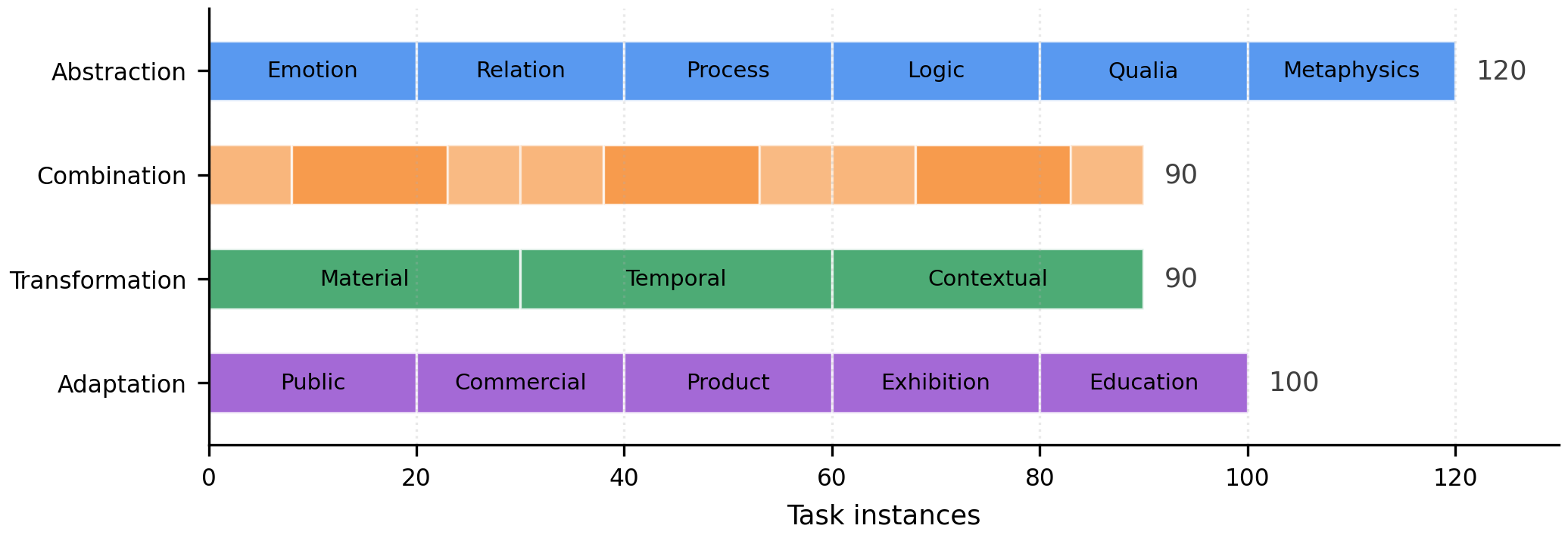}
\caption{Task-family and subtype distribution in the 400-instance \benchmarkname{} task suite.}
\label{fig:app-dataset-taxonomy}
\end{figure*}

\subsection{Subtype Distribution}
Table~\ref{tab:app-subtype-counts} gives the subtype grid used to
balance the benchmark. The combination grid crosses concept type
(\textsc{AA}, \textsc{AC}, \textsc{CC}) with semantic distance
(near, medium, far), where \textsc{AA} denotes abstract--abstract,
\textsc{AC} denotes abstract--concrete, and \textsc{CC} denotes
concrete--concrete.

\begin{table*}[!t]
\centering
\caption{Subtype distribution used for controlled elicitation.}
\label{tab:app-subtype-counts}
\small
\begin{tabularx}{\linewidth}{@{}lYc@{}}
\toprule
\textbf{Family} & \textbf{Subtype grid} & \textbf{Count} \\
\midrule
Abstraction & Emotional/psychological, perceptual/qualia, process/temporal, metaphysical, relational/structural, mathematical/logical & 20 each \\
Combination & \textsc{AA}-near, \textsc{AC}-near, \textsc{CC}-near & 8, 8, 8 \\
Combination & \textsc{AA}-medium, \textsc{AC}-medium, \textsc{CC}-medium & 15 each \\
Combination & \textsc{AA}-far, \textsc{AC}-far, \textsc{CC}-far & 7 each \\
Transformation & Material, temporal, contextual & 30 each \\
Adaptation & Public-interest campaign, commercial communication, digital product, exhibition/public art, educational communication & 20 each \\
\bottomrule
\end{tabularx}
\end{table*}

\begin{figure*}[!t]
\centering
\includegraphics[width=0.94\linewidth]{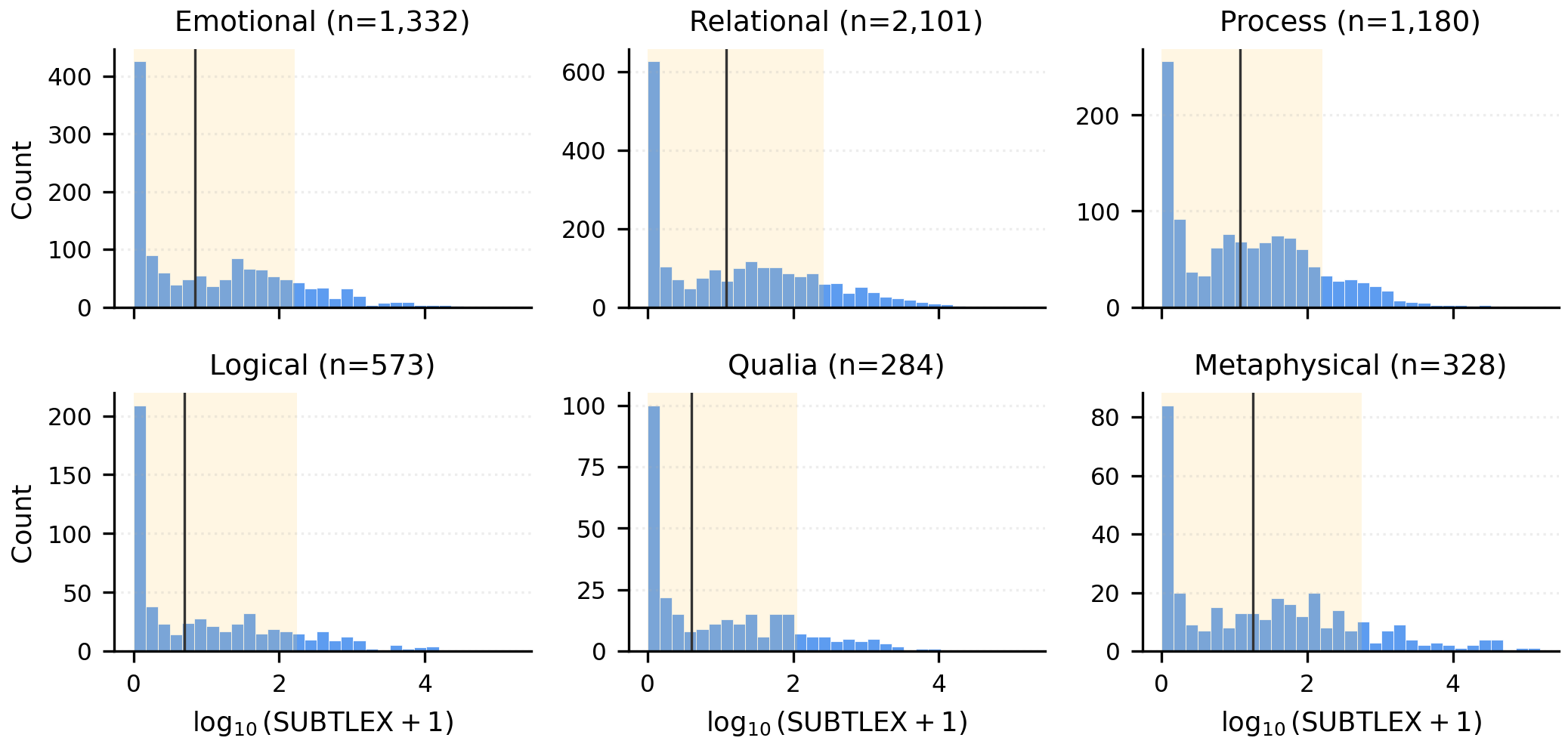}
\caption{Abstraction source-pool diagnostic. Each panel shows the
log-scaled SUBTLEX frequency distribution for nouns in one abstraction
subtype; the shaded interval marks the within-subtype frequency belt
used to avoid extremely rare or ubiquitous lexical items.}
\label{fig:app-abstraction-source-pool}
\end{figure*}

\begin{figure*}[!t]
\centering
\includegraphics[width=0.94\linewidth]{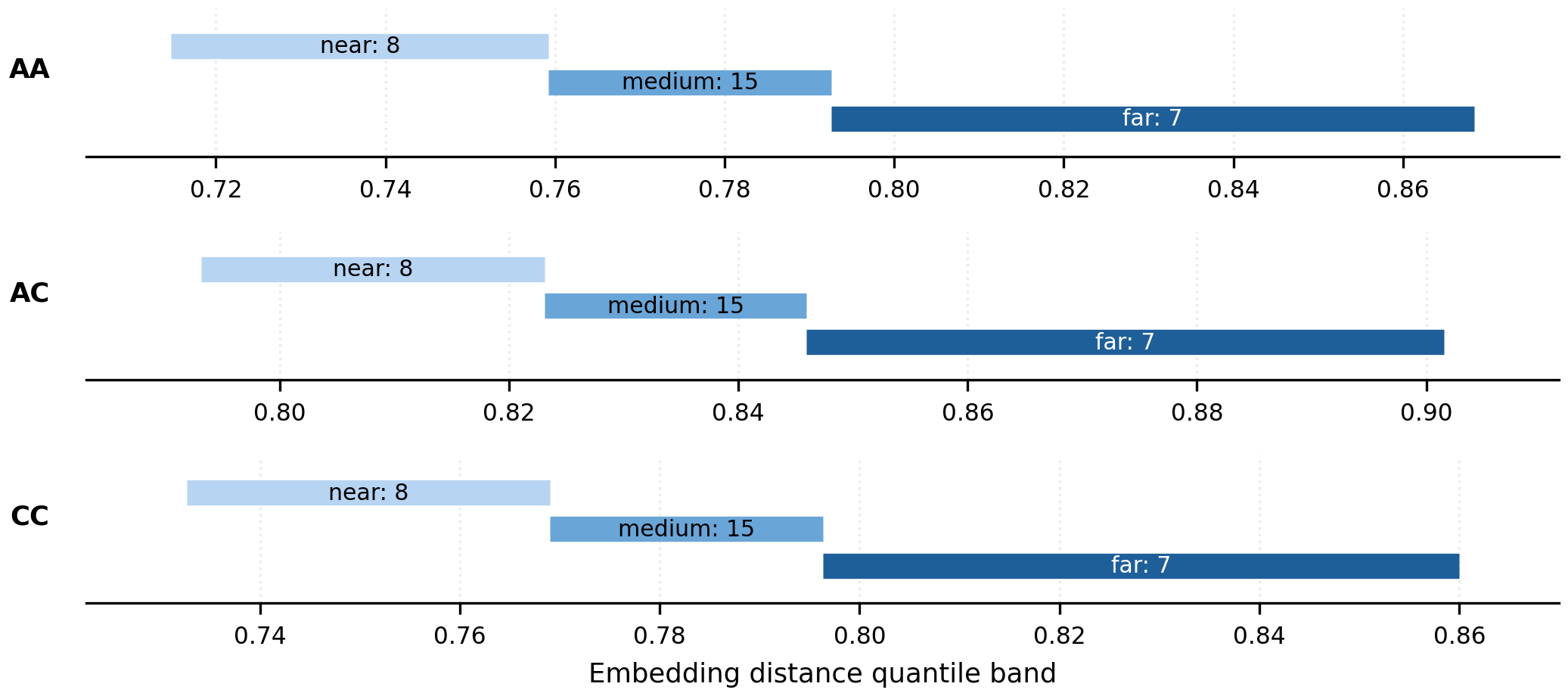}
\caption{Combination sampling diagnostic. Concept pairs are stratified
by pair type and embedding-distance band; numbers inside bands indicate
the selected task count for each cell.}
\label{fig:app-combination-distance-sampling}
\end{figure*}

\begin{figure*}[!t]
\centering
\includegraphics[width=0.94\linewidth]{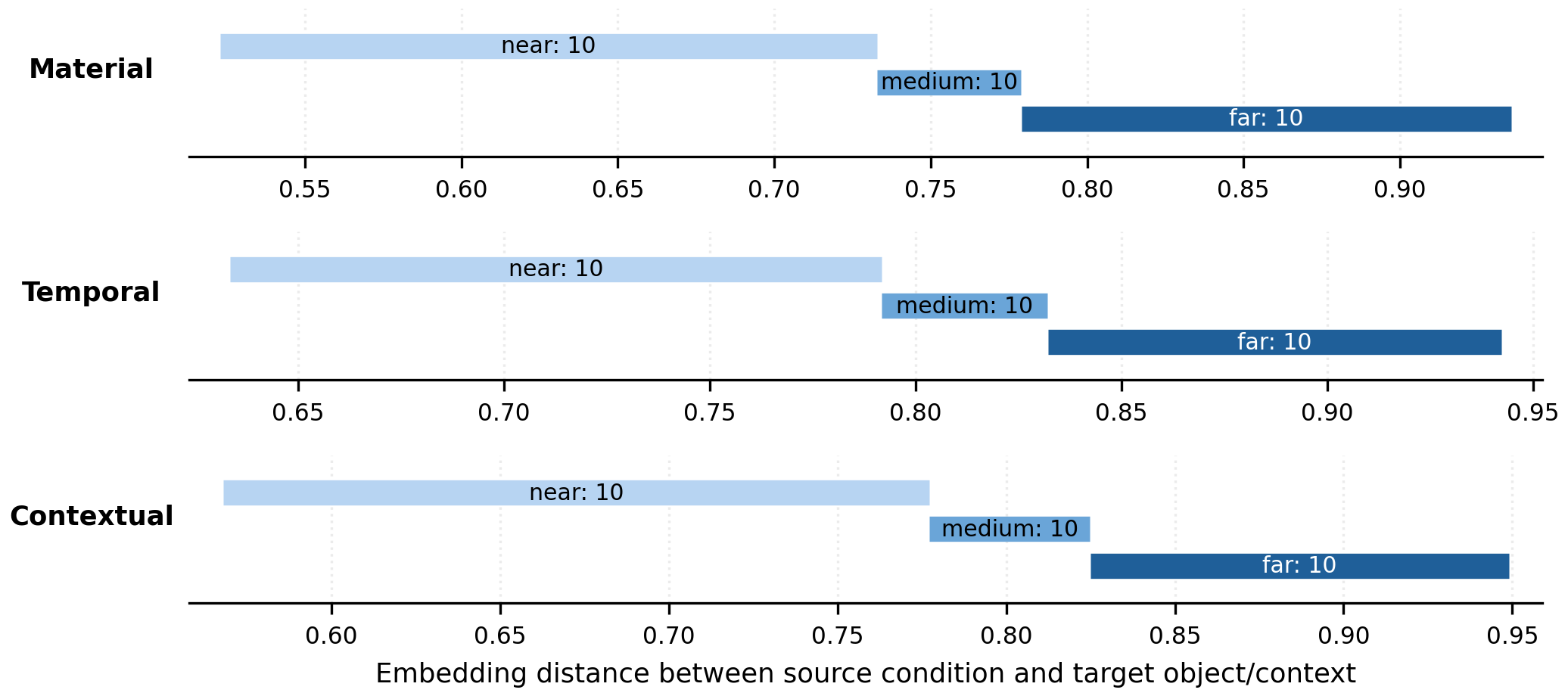}
\caption{Transformation sampling diagnostic. Material, temporal, and
contextual transformations each contribute 30 tasks, with equal counts
from near, medium, and far percentile buckets.}
\label{fig:app-transformation-sampling}
\end{figure*}

Figures~\ref{fig:app-abstraction-source-pool}--\ref{fig:app-transformation-sampling}
give construction diagnostics for the three families whose sampling
depends on lexical or embedding-space controls. These diagnostics are
included to document the elicitation design rather than to report model
performance.

\subsection{Prompt Examples}
Table~\ref{tab:app-prompt-examples} shows compact examples. These
examples are illustrative; full task JSON files store the same fields
using stable ids, family labels, subtype labels, and task properties.

\begin{table*}[!t]
\centering
\caption{Representative task prompts and intended diagnostic use.}
\label{tab:app-prompt-examples}
\small
\begin{tabularx}{\linewidth}{@{}lYY@{}}
\toprule
\textbf{Family} & \textbf{Example brief} & \textbf{What the task tests} \\
\midrule
Abstraction & Express ``time passing'' for a meditation-app launch image while avoiding clock imagery. & Whether the model finds a concrete carrier for an abstract temporal concept. \\
Combination & Fuse coffee culture with Chinese landscape painting and modern minimal design for a vertical brand poster. & Whether sources become one integrated visual system rather than a collage. \\
Transformation & Present a familiar object as if its material, temporal state, or context rewrites its form. & Whether the transformation changes object identity and structure. \\
Adaptation & Create an applied visual deliverable under audience, domain, and medium constraints. & Whether the model chooses a viable strategy under real design constraints. \\
\bottomrule
\end{tabularx}
\end{table*}

\section{Model List and Generation Details}
\label{app:generation_details}
This section reports the model and generation protocol used by the
paper-level evaluation. Internal scripts support smaller pilot runs, but
the appendix follows the main-paper roster and does not treat pilot-only
outputs as final experimental results.

\subsection{Model Selection}
Subject models are selected to cover proprietary frontier systems,
strong open or open-weight systems, and efficient baselines from several
model families. Models used as evaluation judges are excluded from the
subject-generation roster and are listed in the generation configuration
summary. Table~\ref{tab:app-model-roster} lists the display roster used
for reporting. Provider-specific ids are stored in configuration files
and normalized to these display names for the paper.

\begin{table*}[!t]
\centering
\caption{Subject model roster for the official VCI leaderboard.}
\label{tab:app-model-roster}
\small
\begin{tabularx}{\linewidth}{@{}rllY@{}}
\toprule
\textbf{\#} & \textbf{Display model} & \textbf{Role} & \textbf{Notes} \\
\midrule
1 & GPT-5.4 & subject & Frontier OpenAI model. \\
2 & Claude Opus 4.7 & subject & High-end Claude model. \\
3 & Gemini 3.1 Pro Preview & subject & Strong Gemini-family model. \\
4 & Grok 4.3 & subject & Proprietary non-OpenAI comparison. \\
5 & DeepSeek V4 Pro & subject & Strong low-cost reasoning/text model. \\
6 & Qwen3.6 Max Preview & subject & Alibaba/Qwen flagship-style model. \\
7 & GLM 5.1 & subject & Chinese commercial model family. \\
8 & Kimi K2.6 & subject & Long-context Chinese commercial model family. \\
9 & Llama 4 Maverick & subject & Strong open-weight Meta model. \\
10 & MiniMax M2.7 & subject & Commercial model family with reasoning support. \\
11 & DeepSeek R1 & subject & Reasoning-oriented DeepSeek model. \\
12 & Qwen3-8B & subject & Small open-weight Qwen baseline. \\
13 & Phi-4 & subject & Compact model baseline. \\
14 & Llama 3.1 8B Instruct & subject & Small open-weight Meta baseline. \\
\bottomrule
\end{tabularx}
\end{table*}

\subsection{Decoding Parameters}
The implementation uses provider-specific configuration through
OpenRouter-compatible model entries. The common generation defaults are
shown in Table~\ref{tab:app-generation-config}. When a provider exposes
reasoning controls, the configuration records whether reasoning is
enabled, but the output format remains identical across models.

\begin{table}[t]
\centering
\caption{Generation and evaluation configuration summary.}
\label{tab:app-generation-config}
\small
\begin{tabularx}{\linewidth}{@{}L{0.36\linewidth}Y@{}}
\toprule
\textbf{Parameter} & \textbf{Default or policy} \\
\midrule
Provider route & OpenRouter-compatible chat API \\
Temperature & 0.7 for subject generation \\
Maximum tokens & 4096 or 8192 by model family \\
Samples per model--task & 3 candidate samples; sample 1 is the official primary output \\
Official leaderboard output & 1 pre-registered primary response per model--task \\
Typed Idea Graph pool & All 3 samples per model--task \\
Primary judge model & DeepSeek-V4-Flash route \\
Consistency judges & GPT-5.4-mini and Claude-Sonnet-4.6 on fixed sampled pairs \\
Auxiliary sampling & Fixed-count sampled pairs; not proportional to the full pairwise pool \\
JSON repair model & DeepSeek-V4-Flash route \\
Embedding model & text-embedding-3-large route \\
Random seed & 42 for sampling and BT resampling \\
\bottomrule
\end{tabularx}
\end{table}

\subsection{Response Normalization}
Raw model outputs are normalized before evaluation. The extraction
stage uses an LLM-based splitter that returns strict JSON with two keys:
\texttt{idea} and \texttt{visual\_description}. The extractor is
instructed to copy wording faithfully, avoid stylistic rewriting, and
set missing fields to \texttt{null}. The normalized record stores the
source model id, task id, raw-output hash, extraction model, extraction
attempt count, and timestamp.

\begin{methodbox}[breakable=false]
\textbf{Normalization invariant.} Usefulness and Novelty are evaluated
primarily from the extracted \texttt{idea} field, while Expressiveness
is evaluated from \texttt{visual\_description}. This prevents vivid
scene prose from inflating task-fit judgments and prevents brief-fit
reasoning from contaminating expressiveness scores.
\end{methodbox}

\subsection{Extracted Output Length Diagnostics}
\label{app:output-length-diagnostics}
\begin{figure*}[!t]
\centering
\includegraphics[width=0.98\linewidth]{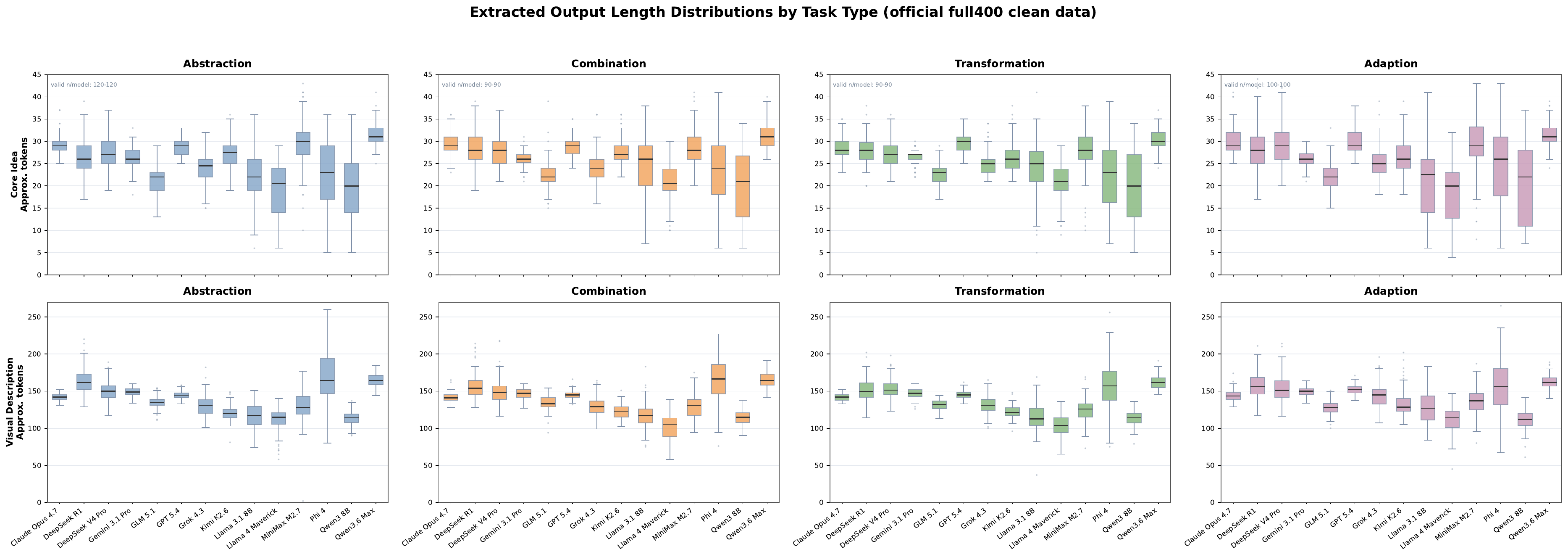}
\caption{Extracted output length distributions by task type for the
official clean dataset. Each box summarizes one subject model within a
task family. The top row reports extracted Core Idea length; the bottom
row reports extracted Visual Description length. This figure is a
format diagnostic rather than a performance result.}
\label{fig:app-extracted-output-lengths}
\end{figure*}

Figure~\ref{fig:app-extracted-output-lengths} reports a lightweight
format diagnostic for the official extracted responses. The diagnostic
uses the 5,600 primary responses in the clean release
(14 models $\times$ 400 tasks), after response normalization and
weak-model extraction audit. The top row shows the length distribution
of the extracted Core Idea field; the bottom row shows the corresponding
Visual Description field. Across task families, Core Ideas are
concentrated around the requested one-sentence length, while Visual
Descriptions cluster around the requested 100--140 word range with
model-specific variation.

\section{Experimental Matrix and Validation Budget}
\label{app:experimental_matrix}
Section~\ref{sec:evaluation} reports the conceptual structure of the
evaluation; this appendix fixes the operational scale used by the final
automatic evaluation. The key design
choice is that three samples are generated for each model--task pair,
but only one pre-registered primary sample enters the official
leaderboard. The remaining samples are used for population-clich\'e
discovery in Typed Idea Graphs, not for best-of-three selection. Because
the official scorer uses active BT rather than exhaustive pairwise
comparison, leave-two-models-out Novelty profiles are built lazily for
the model pairs that are actually selected by the Novelty sampler.

\begin{table*}[!t]
\centering
\caption{Operational scale of the final automatic run.}
\label{tab:app-experiment-matrix}
\small
\begin{tabularx}{\linewidth}{@{}p{0.18\linewidth}p{0.36\linewidth}p{0.22\linewidth}Y@{}}
\toprule
\textbf{Layer} & \textbf{Purpose} & \textbf{Scale} & \textbf{Reported evidence} \\
\midrule
Subject generation & Produce comparable visual plans and a larger population pool for clich\'e discovery. & 14 models $\times$ 400 tasks $\times$ 3 samples = 16,800 outputs & Official sample id, output hashes, normalization logs \\
Official leaderboard & Score one fixed response per model--task pair. & 14 models $\times$ 400 tasks = 5,600 primary responses & Overall and dimension-level BT scores \\
Typed Idea Graphs & Build task-specific population clich\'e profiles for Novelty. & 400 task graphs; 42 ideas per task before leave-out masking & Clich\'e clusters, checklist items, stability diagnostics \\
Lazy leave-two-models-out Novelty & Avoid letting either candidate model define the clich\'es used to judge it. & Built on demand for Novelty comparisons selected by active BT; expected $\leq$ 10,000 contexts & Novelty checklist contexts and TIG audit logs \\
Primary active-BT judge & Run scalable checklist-assisted A/B comparison. & 400 tasks $\times$ 3 dimensions $\times$ 50 judge calls = 60,000 calls & BT estimates, confidence intervals, rank tiers \\
Order-swap diagnostics & Detect A/B position effects. & Approximately 10\% sampled model pairs, about 12,000 calls & Position bias and swap consistency \\
Auxiliary judge consistency & Test whether rankings depend on DeepSeek-V4-Flash. & 2 auxiliary judges $\times$ 3 dimensions $\times$ 100 fixed sampled pairs = 600 calls & Agreement, rank correlation, tie-rate differences \\
Duplicate doublecheck diagnostics & Estimate within-judge repeat stability on sampled pairs. & 2 dimensions $\times$ 200 fixed sampled pairs $\times$ 2 repeated judgments = 800 calls & Duplicate consistency, tie rates, JSON repair rates \\
Novelty ablation & Compare TIG-anchored Novelty with a generic Novelty checklist on the full text-level Novelty validation slice. & 240 paired Novelty cases; 235 with human-majority labels; 240 matched TIG judgments plus 240 no-TIG reruns & Human alignment and checklist specificity \\
Optional construct probes & Check synthetic failure cases such as fluent-but-shallow or clich\'e-template responses. & Planned subset only if automated diagnostics suggest a confound & Probe-specific score behavior \\
\midrule
Completed pairwise judge budget & Includes primary, swap, auxiliary consistency, and duplicate doublecheck calls. & 73,400 completed judge calls before repair overhead & Budget accounting and reliability diagnostics \\
Approximate total chat-call budget & Adds subject generation, atom extraction, canonicalization, and lazy L2MO profile summaries. & About 118,000 calls, excluding embeddings & Budget planning only \\
\bottomrule
\end{tabularx}
\end{table*}

\begin{table*}[!t]
\centering
\caption{Human validation budget. Human judgments support calibration,
reliability analysis, and cross-modal validation for the scalable
evaluation protocol.}
\label{tab:app-human-budget}
\small
\begin{tabularx}{\linewidth}{@{}lYr@{}}
\toprule
\textbf{Human validation component} & \textbf{Design} & \textbf{Judgments} \\
\midrule
Text-level validation & 40 tasks $\times$ 6 pairs $\times$ 3 dimensions $\times$ 3 annotators & 2,160 \\
TIG clich\'e checklist audit & 300 checklist items $\times$ 3 annotators & 900 \\
Faithfulness gate & 40 tasks $\times$ 6 models $\times$ 1 image $\times$ 3 annotators & 720 \\
Image-level preference & 40 tasks $\times$ 4 pairs $\times$ 1 overall judgment $\times$ 3 annotators & 480 \\
\midrule
\textbf{Total} & & \textbf{4,260} \\
\bottomrule
\end{tabularx}
\end{table*}

\subsection{Novelty Ablation Details}
\label{app:novelty_ablation_design}
The completed ablation uses the full text-level Novelty validation
slice: 240 paired Novelty cases from the 40-task validation package. Of these, 235 cases have a final human-majority label. Both conditions use
DeepSeek-V4-Flash as judge and preserve the original response order. The
TIG condition uses the existing case-specific Novelty checklist items
materialized from leave-two-models-out Typed Idea Graphs; the no-TIG
condition reruns the same rows after replacing those task-population
anchors with generic Novelty anchors. All no-TIG judge calls parsed
successfully. The main readout is reported in
Table~\ref{tab:novelty-ablation}.

\subsection{Detailed Reliability Diagnostics}
\label{app:reliability_details}
The main text reports only a compact reliability summary. Tables
\ref{tab:app-reliability-format-order} and
\ref{tab:app-reliability-auxiliary} give the dimension-wise diagnostics
used to support that summary.

\begin{table*}[!t]
\centering
\caption{Format and order diagnostics by dimension.}
\label{tab:app-reliability-format-order}
\small
\begin{tabularx}{\linewidth}{@{}p{0.25\linewidth}cccY@{}}
\toprule
\textbf{Diagnostic} & \textbf{U} & \textbf{E} & \textbf{N} & \textbf{Interpretation} \\
\midrule
Primary JSON repair rate & 0.41\% & 1.01\% & 1.36\% & Low parser intervention \\
Primary tie rate & 13.26\% & 9.90\% & 10.28\% & Non-degenerate tie behavior \\
A-position bias, decisive & +9.9 pp & +11.8 pp & +3.9 pp & Detectable, strongest for Expressiveness \\
Swap decisive consistency & 0.820 & 0.801 & 0.888 & Stable enough for ranking diagnostics \\
Primary--swap BT rank corr. & 0.974 & 0.991 & 0.991 & Leaderboard stable under order reversal \\
\bottomrule
\end{tabularx}
\end{table*}

\begin{table*}[!t]
\centering
\caption{Auxiliary, duplicate, and length diagnostics by dimension.}
\label{tab:app-reliability-auxiliary}
\small
\begin{tabularx}{\linewidth}{@{}p{0.25\linewidth}cccY@{}}
\toprule
\textbf{Diagnostic} & \textbf{U} & \textbf{E} & \textbf{N} & \textbf{Interpretation} \\
\midrule
Winner-longer rate & 0.591 & 0.668 & 0.564 & Verbosity association, strongest for Expressiveness \\
Duplicate consistency & n/a & 0.675 & 0.765 & Pair-level repeatability is higher for Novelty than Expressiveness \\
Aux rank corr. (GPT-5.4-mini) & 0.516 & 0.763 & 0.727 & Moderate to strong auxiliary agreement \\
Aux rank corr. (Claude-Sonnet-4.6) & 0.582 & 0.793 & 0.780 & Stronger auxiliary agreement \\
Aux JSON parse rate & 100\% & 100\% & 100\% & No auxiliary repair required \\
\bottomrule
\end{tabularx}
\end{table*}

\subsection{LLM-Only Pilot and Scaling Criteria}
\label{app:pilot_protocol}
The full evaluation is preceded by two API-facing pilot runs. The goal
is to test the measurement machinery, not to report final model
rankings. All API-facing stages are run sequentially by stage but with
a global concurrency cap of 100 workers.

\begin{table*}[!t]
\centering
\caption{LLM-only pilot scale before the full evaluation. The smoke
stage may use a smaller pairwise budget when it is used only for format
and parser checks.}
\label{tab:app-pilot-scale}
\small
\begin{tabularx}{\linewidth}{@{}lYYY@{}}
\toprule
\textbf{Pilot stage} & \textbf{Generation} & \textbf{Primary judge calls} & \textbf{Additional checks} \\
\midrule
Smoke test & 8 tasks $\times$ 4 models $\times$ 3 samples = 96 outputs & Small sampled or capped pairwise budget & Format, extraction, L2MO profiles, JSON validity \\
Formal active-BT pilot & 20 tasks $\times$ 14 models $\times$ 3 samples = 840 outputs & 20 tasks $\times$ 3 dimensions $\times$ 50-call budget = 3,000 & 10\% order swap; two auxiliary judges on fixed sampled pairs \\
\bottomrule
\end{tabularx}
\end{table*}

The pilot is considered launch-ready only if (i) Core Idea and Visual
Description parsing exceeds 95\%, (ii) L2MO clich\'e profiles are
available and readable for at least 95\% of active-selected Novelty
comparisons, (iii)
judge JSON repair remains below 2\%, (iv) tie rates remain in a
reasonable range rather than collapsing to forced A/B choices, (v)
order-swap position effects are small, (vi) length and adjective-density
correlations do not dominate scores, and (vii) auxiliary judges produce
non-degenerate rank correlations. Human validation and cross-modal
validation are launched only after these pilot gates pass.

\section{Evaluation Prompts}
\label{app:evaluation_details}
The scalable evaluator uses hidden A/B pairwise comparison. For every
comparison, the judge receives a fixed dimension definition, a
dimension-specific evaluation scope, task context when relevant, the
checklist items, and Response A/Response B. The judge must return strict
JSON.

\begin{definitionbox}[breakable=false]
\textbf{Judge output schema.}
{\small\ttfamily\raggedright
\{\\
\quad "dimension":\\
\qquad "usefulness|expressiveness|novelty",\\
\quad "winner": "A|B|tie",\\
\quad "confidence": 0.0,\\
\quad "overall\_rationale":\\
\qquad "one or two sentences",\\
\quad "per\_item\_outcomes": [\\
\qquad \{\\
\qquad\quad "item\_index": 1,\\
\qquad\quad "item": "...",\\
\qquad\quad "winner": "A|B|tie",\\
\qquad\quad "confidence": 0.0,\\
\qquad\quad "evidence\_a": "... or null",\\
\qquad\quad "evidence\_b": "... or null",\\
\qquad\quad "rationale": "..."\\
\qquad \}\\
\quad ]\\
\}
}
\end{definitionbox}

\subsection{Usefulness Checklist}
Usefulness measures whether the Core Idea maps specifically to the task
brief. The prompt explicitly instructs the judge not to score novelty or
visual density. Checklist generation uses three fixed slots: brief fit,
operation completion, and constraint-context fit. A typical item has the
form:
\begin{quote}
\small
\textit{[Inspects: Core Idea] Which response maps more specifically to
the required brief, rather than offering a generic visual theme?}
\end{quote}

The usefulness judge receives the task brief because task fit cannot be
evaluated without knowing the requested concept, sources, audience,
medium, or constraints.

\subsection{Expressiveness Checklist}
Expressiveness measures whether the Visual Description gives enough
specific visual information that two readers would imagine nearly the
same image. It is not a task-fit score and does not reward novelty. The
prompt includes task context before generating checklist items: in
Combination it highlights the fused entity, in Transformation the
transformed object, and in Adaptation the entire intended deliverable.
Checklist generation then uses three fixed slots: visual grounding,
spatial organization, and appearance specificity. A typical item has the
form:
\begin{quote}
\small
\textit{[Inspects: Visual Description] Which description shows more of
the actual scene, instead of explaining what it means?}
\end{quote}

The expressiveness judge is instructed to compare only the response's
own visual language. This separation is important because an expressive
description can still be useless for the brief, and a useful concept can
be visually under-specified.

\subsection{Novelty Checklist}
Novelty measures whether a response avoids obvious, stock, or
population-common visual ideas for the same task. The prompt uses the
brief only as context for what would be obvious. The first checklist
item is replaced by a task-specific item materialized from the Typed
Idea Graph profile. Checklist generation uses three fixed slots:
clich\'e distance, source-domain shift, and perspective or relation
shift. This keeps Novelty separate from brief fit and visual vividness.

\section{Bradley--Terry Aggregation Details}
Pairwise preferences are aggregated with a Bradley--Terry model. Let
$\theta_m$ denote the latent skill of model $m$ for one dimension. The
core quantities are summarized below.

\begin{figure*}[!t]
\centering
\begin{minipage}{0.94\linewidth}
\small
\textbf{Bradley--Terry aggregation summary.}
\begin{align*}
P(i \succ j)
&=\sigma(\theta_i-\theta_j)
=\frac{1}{1+\exp[-(\theta_i-\theta_j)]},\\
\mathcal{L}(\theta)
&=-\sum_{r\in\mathcal{D}} w_r\log\sigma(\theta_{a_r}-\theta_{b_r})
+\lambda\sum_m\theta_m^2,\qquad
\sum_m\theta_m=0,\\
s(i,j)
&=\left(\Sigma_{ii}+\Sigma_{jj}-2\Sigma_{ij}\right)(1+n_{ij})^{-1/2},\\
\widehat{W}_m
&=\frac{W_m+0.5T_m}{W_m+L_m+T_m}.
\end{align*}
\end{minipage}
\end{figure*}

Here $w_r$ is judge confidence, $a_r$ and $b_r$ are the preferred and
non-preferred models, $s(i,j)$ is the active-pair score, and
$\widehat{W}_m$ is the diagnostic win rate with ties as half-wins.

\subsection{Pairwise Sampling}
For each task, the system first covers the model pool with warm-start
pairs and then allocates remaining budget to uncertain active pairs.

\begin{algorithm*}[!t]
\caption{Checklist-assisted pairwise BT evaluation}
\label{alg:pairwise-bt}
\begin{algorithmic}[1]
\Require Tasks $\mathcal{T}$, models $\mathcal{M}$, dimension $d$, budget $B$
\Ensure Leaderboard for dimension $d$
\State $\mathcal{R}\gets\emptyset$
\ForAll{$t\in\mathcal{T}$}
  \State Select warm-start and active pairs under budget $B$
  \State Randomize A/B order and query checklist judge for dimension $d$
  \State Add decisive preferences to $\mathcal{R}$; retain ties for diagnostics
\EndFor
\State Fit BT skills $\hat{\theta}$ from $\mathcal{R}$
\State Bootstrap tasks to estimate CIs and rank probabilities
\State \Return dimension leaderboard from $\hat{\theta}$
\end{algorithmic}
\end{algorithm*}

\subsection{Tie Handling}
The judge may return \texttt{tie}. Ties are retained for diagnostics,
tie rate, raw win rate, Borda-style summaries, and tie-as-half-win
diagnostic win rates. Only decisive comparisons enter the main BT
likelihood. If a response is empty, an automatic comparison record is
generated: non-empty responses beat empty responses with confidence
1.0, and two empty responses tie.

\subsection{Confidence Intervals}
Task-level bootstrap intervals are computed by resampling tasks with
replacement, refitting BT on the sampled comparisons, and taking
percentiles of $\theta_m$. The reported appendix intervals use the
5th, 50th, and 95th percentiles unless otherwise stated. Rank
probability is estimated by drawing
\[
\tilde{\theta}\sim \mathcal{N}(\hat{\theta},\hat{\Sigma})
\]
and counting the frequency with which each model occupies each rank.

\subsection{Rank Stability}
Rank stability is reported using bootstrap intervals, rank probability
mass, and pairwise indistinguishability tiers. Two neighboring models
are treated as indistinguishable when their BT difference is smaller
than the corresponding Wald interval.

\section{Typed Idea Graphs for Population-Anchored Novelty}
\label{app:tig_details}
Typed Idea Graphs operationalize population-anchored novelty. The goal
is not to ask whether a response is far from a generic embedding
centroid, but whether it avoids the specific ideas that many LLMs
converge on for the same task.

\subsection{Atom Schema}
For each Core Idea, the extractor returns 3--7 typed atoms:

\begin{table}[t]
\centering
\caption{Typed idea atom schema.}
\label{tab:app-atom-schema}
\small
\begin{tabularx}{\linewidth}{@{}L{0.25\linewidth}Y@{}}
\toprule
\textbf{Type} & \textbf{Meaning} \\
\midrule
concept & Central solution or representational strategy. \\
motif & Visual or semantic element used by the idea. \\
mechanism & Technique, transformation, or implementation move. \\
purpose & Intended effect, meaning, or communicative function. \\
audience & Target audience or user group, only if explicit. \\
\bottomrule
\end{tabularx}
\end{table}

Typed edges connect atoms within a response, such as
\texttt{motif supports purpose} or \texttt{mechanism realizes concept}.

\subsection{Extraction and Canonicalization Prompt}
The extraction prompt takes the task brief and Core Idea as input. It
instructs the model to avoid extracting lighting, color, texture, or
composition unless those properties are central to the idea. The
canonicalization prompt groups synonymous or near-synonymous atoms
within each type, requiring every input atom to appear in exactly one
alias list and prohibiting invented atoms.

For an atom $u$, the canonical record stores id, type, canonical label,
aliases, provenance, and optionally an embedding vector. Provenance links
the atom back to response id, model id, task id, and sample index.

\subsection{Graph Edge Definitions}
The graph contains co-occurrence, same-type similarity, and extracted
typed-relation edges. The scoring quantities are:

\begin{figure*}[!t]
\centering
\begin{minipage}{0.94\linewidth}
\small
\textbf{Typed Idea Graph scoring summary.}
\begin{align*}
w_{\mathrm{cooc}}(u,v)
&=\max\left(0,\frac{\log \frac{p(u,v)}{p(u)p(v)}}{-\log p(u,v)}\right),\\
w_{\mathrm{sim}}(u,v)
&=\cos(e_u,e_v)\cdot \mathbf{1}[\cos(e_u,e_v)\geq 0.78],\\
w_{\mathrm{typed}}(u,v,r)
&=\sum_{\ell}\mathbf{1}[(u,v,r)\in E_{\ell}],\\
\mathrm{ModelCov}(C)
&=\frac{|\{m:\exists u\in C,\;m\in \mathrm{prov}(u)\}|}{|\mathcal{M}|},\\
\mathrm{Cent}(C)
&=\frac{\sum_{u\in C}\deg_w(u)}{|C|\max_{u'}\deg_w(u')},\qquad
\mathrm{Conv}(C)=
\frac{2}{|C|(|C|-1)}\sum_{u<v\in C}\max(0,\cos(e_u,e_v)),\\
\mathrm{ClicheScore}(C)
&=0.45\,\mathrm{ModelCov}(C)
+0.35\,\mathrm{Cent}(C)
+0.20\,\mathrm{Conv}(C).
\end{align*}
\end{minipage}
\end{figure*}

\subsection{Clich\'eScore Computation}
Clusters are formed within same-type subgraphs using hierarchical
Leiden when available and weighted connected components otherwise.
High-scoring clusters are common across models, central in the task's
idea graph, and internally convergent; these clusters become the
population-anchored cliche candidates.

\subsection{Leave-Two-Models-Out Scoring}
\begin{methodbox}
\textbf{Anti-leakage rule.} When comparing candidate models $m_a$ and
$m_b$ on task $t$, the population reference graph is built from
$\mathcal{M}\setminus\{m_a,m_b\}$ and uses all three samples from each
remaining model. With 14 subject models, each pairwise comparison uses
36 reference ideas. Neither candidate model is allowed to define the
clich\'es used to judge the pair.
\end{methodbox}

\begin{algorithm*}[!b]
\caption{Leave-two-models-out Typed Idea Graph novelty}
\label{alg:ltmo-tig}
\begin{algorithmic}[1]
\Require Task $t$, model population $\mathcal{M}$, sample pool $Y_t$
\Ensure Novelty checklist for each judged model pair
\ForAll{judged pair $q=\{m_a,m_b\}\subset\mathcal{M}$}
  \State $Y^{-q}_t\gets\{y_{m',t,s}:m'\notin q,\;s\in\{1,2,3\}\}$
  \State Extract and canonicalize typed atoms from $Y^{-q}_t$
  \State Build graph $G^{-q}_t$ with co-occurrence, similarity, and typed edges
  \State Cluster $G^{-q}_t$ and rank clusters by ClicheScore
  \State Materialize top clusters as novelty checklist items
  \State Score the primary outputs of $m_a$ and $m_b$ with checklist-BT
\EndFor
\end{algorithmic}
\end{algorithm*}

This procedure makes novelty population-anchored but not self-anchored.
It also makes the novelty target task-specific: the reference population
for a coffee-brand poster is different from the reference population for
a temporal object transformation.

\subsection{Checklist Materialization Examples}
Ranked clusters are converted into natural-language checklist items that
ask whether a response avoids, transforms, or merely repeats the common
pattern. For example, if a task's reference graph contains a high-scoring
cluster around ``plants, leaves, soft green growth'' for calmness, the
materialized item can ask:
\begin{quote}
\small
\textit{Which response avoids relying on common plant-growth imagery
unless it transforms that motif into a more specific visual mechanism?}
\end{quote}

\begin{definitionbox}
\textbf{Why not embedding distance alone?} Embedding distance measures
generic semantic separation, but visual clich\'es are task-conditioned.
A sunset silhouette may be semantically appropriate and far from some
generic centroid, yet it is clich\'ed if many models independently use it
for the same prompt. Typed Idea Graphs measure population convergence
within the task, not just semantic distance in embedding space.
\end{definitionbox}

\subsection{TIG Checklist Audit Schema}
The TIG checklist audit is designed to validate the novelty checklist
items themselves before treating them as reliable judge anchors. The
audit does not ask annotators a single vague question such as whether an
item is ``reasonable.'' Instead, each materialized Novelty checklist
item is evaluated along five concrete criteria, using the item text and
its underlying cluster/provenance summary.

\begin{table}[t]
\centering
\caption{TIG Novelty checklist audit schema. Each materialized checklist
item is audited as a population-clich\'e anchor, not as a direct
preference judgment between A and B.}
\label{tab:tig-checklist-audit}
\small
\begin{threeparttable}
\begin{tabularx}{\linewidth}{@{}L{0.30\linewidth}Y@{}}
\toprule
\textbf{Audit item} & \textbf{Question} \\
\midrule
Cluster validity
& Does the item represent a visual pattern that recurs across multiple
models, rather than a one-off phrasing from a single response? \\
Visual specificity
& Is the item a visual clich\'e rather than a language-expression
clich\'e, abstract rhetorical move, or evaluative phrase? \\
Faithfulness
& Does the item faithfully summarize the original idea cluster without
adding objects, mechanisms, or intents not supported by the cluster? \\
Non-leakage
& Does the item avoid leaking content from either A/B candidate response
in the leave-two-models-out comparison? \\
Usefulness for judging
& Does the item help a judge decide whether a candidate repeats,
transforms, or avoids the population pattern? \\
\bottomrule
\end{tabularx}
\begin{tablenotes}[flushleft]\footnotesize
\item All criteria use the same codes: \texttt{pass}, \texttt{partial},
\texttt{fail}, and \texttt{not\_enough\_evidence}.
\end{tablenotes}
\end{threeparttable}
\end{table}

The audit record stores
\texttt{audit\_id}, \texttt{task\_id}, \texttt{comparison\_key},
\texttt{checklist\_item\_id}, \texttt{checklist\_text},
\texttt{cluster\_id}, \texttt{cluster\_summary}, the five criterion
codes in Table~\ref{tab:tig-checklist-audit}, \texttt{overall\_action},
\texttt{revision\_note}, \texttt{annotator\_id}, and timestamp.
\texttt{overall\_action} is one of \texttt{accept}, \texttt{revise},
or \texttt{exclude}. We accept items that pass all criteria or contain
only minor partial issues, revise items whose cluster is valid but whose
wording is underspecified or not sufficiently visual, and exclude items
that fail cluster validity, visual specificity, faithfulness, or
non-leakage. This audit is independent of pairwise A/B preference
annotation; pairwise annotators may flag an item and leave a note if
they observe an audit failure during judging.

\section{Human Annotation Protocol}
Human annotations support calibration and reliability analysis on
targeted validation slices. The main text-level validation subset
contains 40 tasks, stratified as 10
tasks from each family. For each task, we sample six model pairs and ask
annotators to judge Usefulness, Expressiveness, and Novelty separately,
with three annotators per item. This yields 2,160 text-level human
judgments.

\subsection{Annotator Instructions}
Annotators see one task and two anonymized responses. Model identity is
hidden, order is randomized, and a tie option is available. The interface
asks annotators to judge only one dimension at a time:

\begin{itemize}\itemsep0pt
\item \textbf{Usefulness:} Which Core Idea better satisfies the brief,
constraints, audience, medium, and required source concepts?
\item \textbf{Expressiveness:} Which Visual Description gives a more
concrete and stable mental image?
\item \textbf{Novelty:} Which response better avoids task-specific
population clich\'es while remaining meaningful for the brief?
\end{itemize}

Annotators are instructed not to reward verbosity by itself and not to
infer unstated intentions.

\subsection{Pairwise Annotation Guidelines}
The annotation interface exposes one dimension at a time. Annotators
are instructed to treat the checklist as a reference cue rather than a
rubric with equal-weight items. They must not count how many checklist
items each side appears to satisfy and mechanically choose the side with
more item-level wins. The final label is a single overall A/B/tie
preference for the current dimension.

\begin{table*}[!t]
\centering
\caption{Human pairwise annotation guidelines by dimension.}
\label{tab:human-guidelines}
\small
\begin{tabularx}{\linewidth}{@{}p{0.22\linewidth}YY@{}}
\toprule
\textbf{Dimension} & \textbf{Primary field} & \textbf{Decision rule} \\
\midrule
Usefulness
& Core Idea
& Prefer the idea that more specifically satisfies the brief,
constraints, audience, medium, and required visual operation. Do not
reward verbal polish or visual density by itself. \\
Expressiveness
& Visual Description
& Prefer the description that supports a more concrete, stable, and
drawable mental image. Do not reward task fit, novelty, length, or
atmospheric language by itself. \\
Novelty
& Core Idea plus task-specific clich\'e references
& Prefer the idea that better avoids or transforms population-common
visual patterns while remaining meaningful for the brief. Do not reward
randomness, obscurity, or off-brief weirdness. \\
\bottomrule
\end{tabularx}
\end{table*}

Ties are allowed only when the current dimension has no stable,
explainable winner: both responses are similarly strong, similarly weak,
or their advantages cancel out within the current dimension. Annotators
are explicitly told not to use tie as a substitute for uncertainty when
one response has a clear dimension-specific advantage.

\subsection{Flagging and Notes}
The interface includes a flag option for records that require later
review. Flagging is not a fourth preference label and is not a substitute
for tie. When possible, annotators still choose A, B, or tie, and use the
note field to explain why the record was flagged.

Annotators are instructed to flag missing or truncated responses,
field--dimension mismatches, ambiguous briefs, checklist/task mismatch,
translation problems, suspected A/B duplication or model-identity
leakage, violations of the single-static-image protocol, and Novelty
checklist items that appear to fail the TIG audit criteria in
Table~\ref{tab:tig-checklist-audit}. They are instructed not to flag
merely because a case is difficult, because A and B are close, or because
they personally dislike a style. Notes should identify the issue
concisely, for example \texttt{TIG audit: non-leakage concern} or
\texttt{visual field truncated}.

\subsection{Calibration Examples}
Calibration examples cover three common borderline cases: useful but
clich\'ed responses, vivid but task-misaligned responses, and unusual
but under-specified responses. Annotators first inspect the checklist
anchors, then compare A/B responses, and finally select A, B, or tie.
The examples are designed to make the three dimensions separable rather
than to teach annotators a single global notion of quality.

\subsection{Pair Sampling}
Pairs are sampled from preliminary DeepSeek-V4-Flash rankings but shown
to annotators without model identities or scores. For each task, the six
pairs include top-vs-bottom, top-vs-middle, middle-vs-middle, and
dimension-tradeoff comparisons, such as a highly expressive response
against a more novel but less polished response. This makes the
validation subset include both obvious and borderline comparisons.

\subsection{Agreement Metrics}
We report raw agreement, tie rate, human-vs-judge rank correlation, and
pairwise consistency, following standard multi-rater reliability
practice~\citep{fleiss1971measuring,shrout1979intraclass}. If $h_{ij}$
is the human preference sign for model pair $(i,j)$ and $a_{ij}$ is the
automated preference sign, pairwise agreement is
\[
\mathrm{Agree}=
\frac{1}{|\mathcal{P}|}
\sum_{(i,j)\in\mathcal{P}}\mathbf{1}[h_{ij}=a_{ij}],
\]
with ties counted as agreement only when both sides tie. Human-subset
BT rankings are compared to automated BT rankings with Spearman
correlation and Kendall correlation.

\subsection{Completed Human Annotation Reliability}
Table~\ref{tab:human-annotation-reliability} summarizes the completed
human annotation results used in this paper. For the TIG checklist audit,
we report only the final \texttt{accept}/\texttt{revise}/\texttt{exclude}
action label; the five finer-grained TIG audit fields are reserved for
internal revision rather than main reliability reporting.

\begin{table*}[!t]
\centering
\caption{Completed human annotation reliability summary.}
\label{tab:human-annotation-reliability}
\small
\begin{tabularx}{\linewidth}{@{}L{0.21\linewidth}L{0.16\linewidth}L{0.18\linewidth}YY@{}}
\toprule
\textbf{Subset} & \textbf{Items / judgments} & \textbf{Agreement} & \textbf{Consensus} & \textbf{Majority outcome} \\
\midrule
Text-level pairwise validation & 720 / 2,160 & Fleiss $\kappa=0.596$ & 62.5\% unanimous; 97.6\% majority & A=312, B=276, tie=115, no majority=17 \\
TIG checklist audit action & 300 / 900 & Fleiss $\kappa=0.762$ & 78.3\% unanimous; 99.0\% majority & accept=86, revise=153, exclude=58, no majority=3 \\
Faithfulness gate & 240 / 720 & pass/fail $\kappa=0.637$; score ICC=0.711 & 77.9\% unanimous; 100.0\% majority & pass=179, fail=61 \\
Image-level preference & 160 / 480 & Fleiss $\kappa=0.690$ & 73.8\% unanimous; 98.8\% majority & A=74, B=73, tie=11, no majority=2 \\
\bottomrule
\end{tabularx}
\end{table*}

\section{Cross-Modal Grounding Protocol}
Cross-modal grounding tests whether text-level VCI scores reflect
visual content rather than surface language quality. The protocol has
two stages: a faithfulness gate and image-level pairwise preference.

\subsection{Text-to-Image Rendering Setup}
The grounding subset contains 40 tasks, stratified as 10 from each task
family. We select six representative subject models from the
text-level leaderboard: two top models, two middle models, and two
bottom models. Each textual visual plan is converted into a rendering
prompt and passed to a fixed text-to-image system with one seed,
yielding $40\times 6\times 1=240$ images. The rendering prompt is
constructed from the Core Idea and Visual Description, with no model
identity and no evaluation score exposed. This subset is designed as a
targeted grounding check rather than a second full-scale benchmark.

\subsection{Faithfulness Annotation}
Renderings pass the faithfulness gate only if they preserve the source
idea at the level needed for image-level preference. The default rule is
\[
\begin{aligned}
\mathrm{Pass}(I,y)
&=\mathbf{1}\bigl[\mathrm{Faith}(I,y)\geq 4/5\\
&\qquad\land \neg \mathrm{HardFail}(I,y)\bigr].
\end{aligned}
\]
Hard failures include major semantic drift, missing core object,
concept splitting in fusion tasks, object identity loss in
transformation tasks, and style/context mismatch in applied design
tasks.

Each image receives three independent faithfulness annotations, for a
total of 720 judgments. We report pass rates by model and by task
family, the correlation between faithfulness pass rate and text-level
VCI score, and both faithful-only and rendering-failure-penalized
versions of the cross-modal agreement analysis.

\begin{methodbox}
\textbf{Faithfulness gate.} Image-level validation is conducted only on
faithful renderings. This prevents a poor renderer from being mistaken
for poor visual ideation by the text model.
\end{methodbox}

\subsection{Image-Level Preference Annotation}
After filtering, human judges compare images alone. They do not see the
original text, model identity, or text-level ranking. The image-level
criteria mirror the VCI dimensions: task usefulness when task context is
shown, visual expressiveness in the rendered image, and visible novelty
relative to alternatives for the same task.

For each of the 40 tasks, we sample four image pairs after the
faithfulness gate and collect three annotations for one overall visual
preference judgment, yielding 480 image-level preference judgments.
The overall judgment asks which image better realizes a useful,
expressive, and non-clich\'ed visual plan as a whole; dimension-specific
image judgments are reserved for optional follow-up analysis.

\subsection{Agreement and Correlation Metrics}
For each model pair $(m,n)$, let $\Delta^{\mathrm{text}}_{mn}$ be the
text-level BT difference and let $p^{\mathrm{img}}_{mn}$ be the
image-level preference rate for model $m$ over model $n$. The primary
reported readout is pairwise sign agreement, with rank correlations
reserved as exploratory diagnostics rather than headline evidence:
\[
\begin{aligned}
\mathrm{PairAgree}
&=\frac{1}{|\mathcal{P}|}\sum_{(m,n)\in\mathcal{P}}
\mathbf{1}\Bigl[\\
&\quad \operatorname{sign}(\Delta^{\mathrm{text}}_{mn})
=\operatorname{sign}(p^{\mathrm{img}}_{mn}-0.5)
\Bigr].
\end{aligned}
\]
The main text reports the agreement counts rather than emphasizing
aggregate rank correlations, because the six-model subset is a targeted
validation slice rather than a full image-level leaderboard. Cross-modal
agreement is interpreted as convergent validity evidence, not as proof
that text-level evaluation is sufficient on its own.

\section{Adaptation Strategy Taxonomy}
Adaptation tasks are useful for strategy analysis because the prompt
does not prescribe a single creative operation. We therefore use a
multi-label taxonomy to describe how models respond to applied
constraints.

\begin{figure*}[!t]
\centering
\includegraphics[width=0.92\linewidth]{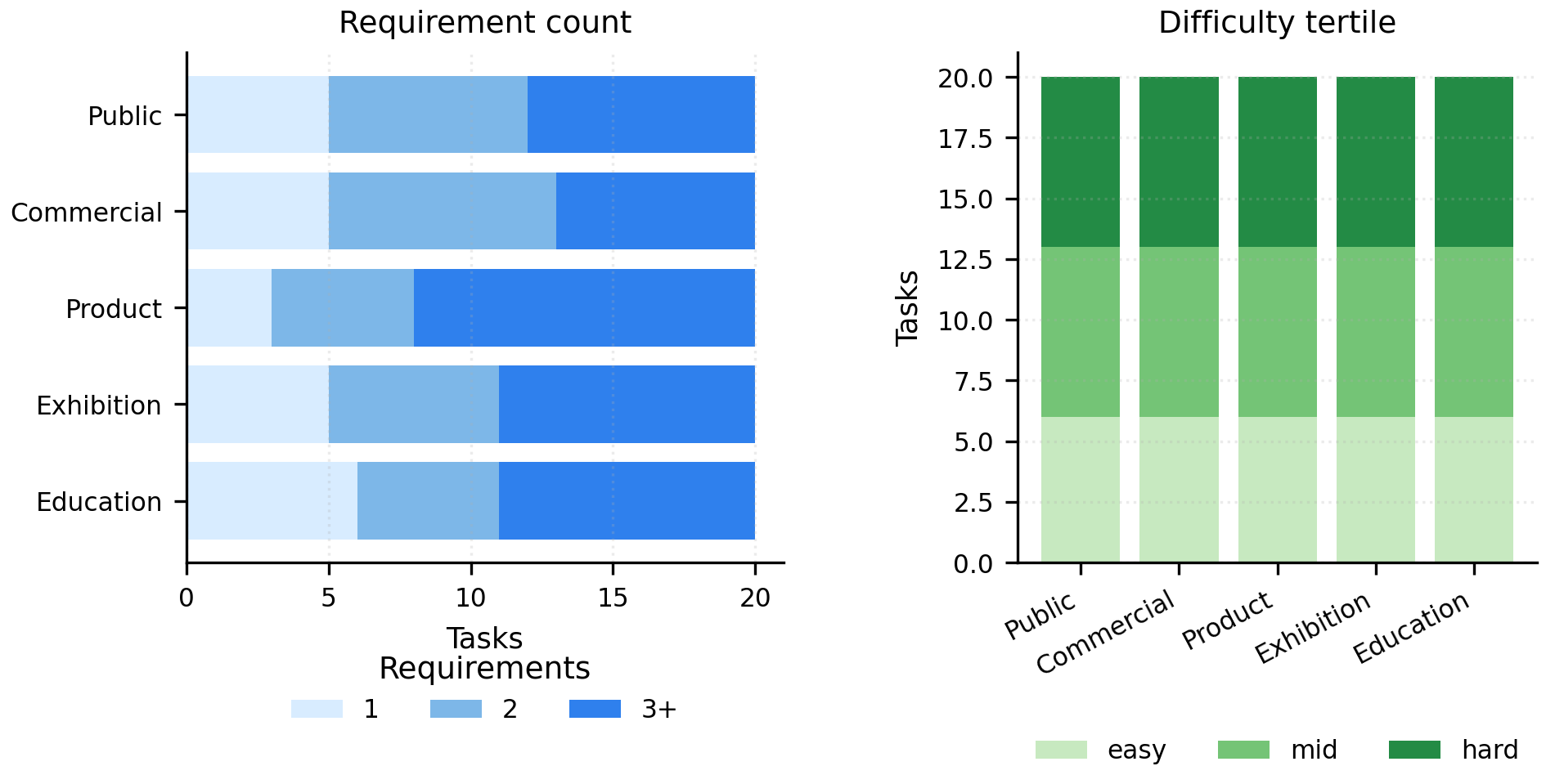}
\caption{Adaptation constraint diagnostic by applied domain.}
\label{fig:app-adaptation-constraint-distribution}
\end{figure*}

Figure~\ref{fig:app-adaptation-constraint-distribution} summarizes the
constraint profile of the released adaptation tasks. The taskset is
balanced across applied domains while retaining variation in the number
of explicit requirements and in the difficulty tertile assigned during
task construction.

\subsection{Strategy Definitions}
\begin{table}[t]
\centering
\caption{Multi-label adaptation strategy taxonomy.}
\label{tab:app-adaptation-taxonomy}
\small
\begin{tabularx}{\linewidth}{@{}L{0.34\linewidth}Y@{}}
\toprule
\textbf{Strategy} & \textbf{Definition} \\
\midrule
Constraint literalization & Directly turns a written constraint into a visible object, label, or layout device. \\
Audience reframing & Changes subject, tone, or visual language to fit the intended audience. \\
Medium adaptation & Uses the affordances of the requested medium, such as poster, app screen, package, or public installation. \\
Metaphor substitution & Replaces an obvious applied-design trope with a new metaphorical carrier. \\
Scenario embedding & Places the message inside a specific scene of use rather than a generic symbolic image. \\
Style transfer & Applies a recognizable art, branding, or cultural style to the solution. \\
Risk-avoidant generic design & Produces a safe but low-specificity design that satisfies surface constraints without a distinctive idea. \\
\bottomrule
\end{tabularx}
\end{table}

\subsection{Positive and Negative Examples}
A positive \textit{medium adaptation} example uses the physical or
interaction constraints of the medium as part of the image concept. A
negative example merely says that the design is ``poster-like'' or
``suitable for an app'' without making the medium visible. A positive
\textit{audience reframing} example changes scale, tone, imagery, or
information density for the intended audience. A negative example only
names the audience in the rationale.

\subsection{Multi-Label Annotation Schema}
Each response may receive zero or more labels. The annotation record is
\[
z_{m,t}=\{s_1,\ldots,s_k\},\quad s_i\in\mathcal{S},
\]
where $\mathcal{S}$ is the strategy set in Table~\ref{tab:app-adaptation-taxonomy}.
Annotators also mark whether the strategy is central or incidental. A
strategy is central if removing it would change the core visual idea;
it is incidental if it appears only as surface styling.

\section{Additional Results}
This section provides appendix-only result formats and diagnostic
templates. These materials support the main argument but are not needed
to understand the benchmark.

\subsection{Full Model Rankings}
Table~\ref{tab:app-full-ranking} gives the final official 14-model
leaderboard. Overall VCI is the mean of the three standardized
dimension scores. Expected winrates are BT-derived pairwise win
probabilities against the model population for each dimension.

\begin{table*}[!t]
\centering
\caption{Full model ranking with standardized BT scores and expected winrates.}
\label{tab:app-full-ranking}
\scriptsize
\setlength{\tabcolsep}{3pt}
\begin{tabular}{@{}r l *{4}{S[table-format=+1.2]} *{3}{S[table-format=1.2]}@{}}
\toprule
\textbf{Rank} & \textbf{Model} &
\multicolumn{1}{c}{\textbf{Overall}} &
\multicolumn{1}{c}{\textbf{U $z$}} &
\multicolumn{1}{c}{\textbf{E $z$}} &
\multicolumn{1}{c}{\textbf{N $z$}} &
\multicolumn{1}{c}{\textbf{U win}} &
\multicolumn{1}{c}{\textbf{E win}} &
\multicolumn{1}{c}{\textbf{N win}} \\
\midrule
1 & Gemini-3.1 Pro & +0.96 & +0.61 & +1.19 & +1.08 & 0.61 & 0.80 & 0.66 \\
2 & GPT-5.4 & +0.92 & +1.53 & +0.65 & +0.60 & 0.78 & 0.66 & 0.59 \\
3 & Claude-4.7 & +0.73 & +0.59 & +1.20 & +0.39 & 0.61 & 0.80 & 0.56 \\
4 & DeepSeek-V4 & +0.70 & +0.49 & +0.70 & +0.90 & 0.59 & 0.68 & 0.64 \\
5 & Kimi-K2.6 & +0.68 & +0.13 & +0.26 & +1.66 & 0.52 & 0.57 & 0.75 \\
6 & Grok-4.3 & +0.49 & +1.10 & +0.55 & -0.17 & 0.70 & 0.64 & 0.47 \\
7 & DeepSeek-R1 & +0.49 & +0.38 & +0.88 & +0.20 & 0.57 & 0.72 & 0.53 \\
8 & GLM-5.1 & +0.39 & +0.43 & +0.44 & +0.31 & 0.58 & 0.61 & 0.55 \\
9 & Qwen3.6-Max & +0.29 & +0.12 & +0.37 & +0.37 & 0.52 & 0.59 & 0.56 \\
10 & MiniMax-M2.7 & -0.15 & +0.45 & -0.37 & -0.52 & 0.58 & 0.41 & 0.42 \\
11 & Qwen3-8B & -1.22 & -1.37 & -1.37 & -0.92 & 0.25 & 0.15 & 0.36 \\
12 & Llama-3.1-8B & -1.23 & -1.95 & -1.46 & -0.28 & 0.14 & 0.13 & 0.46 \\
13 & Phi-4 & -1.42 & -1.46 & -1.57 & -1.24 & 0.23 & 0.11 & 0.31 \\
14 & Llama-4-Mav & -1.64 & -1.06 & -1.46 & -2.39 & 0.31 & 0.13 & 0.14 \\
\bottomrule
\end{tabular}
\end{table*}

\subsection{Full Task-Family Results}
Task-family residuals are reported rather than only raw means. For
dimension $d$ and family $f$, the residual is
\[
R_{f,d}=\bar{s}_{f,d}-\bar{s}_{\cdot,d}.
\]
Positive values indicate that a family is above the dimension average;
negative values indicate that it is below the dimension average. The
main analysis shows that Combination is below average across dimensions,
Transformation is relatively strong for Usefulness and Novelty, and
Abstraction is especially strong for Expressiveness.

\subsection{Additional Failure Cases}
We use the following diagnostic template for failure cases:
\[
\begin{aligned}
\mathrm{Failure}(y)=(&\mathrm{family},\mathrm{dimension},
\mathrm{symptom},\\
&\mathrm{likely\ cause},\mathrm{evidence}).
\end{aligned}
\]
Common symptoms include:
\begin{itemize}\itemsep0pt
\item \textbf{Clich\'e-heavy abstraction:} the response uses a common
symbol such as a sunset, mirror, plant, or broken object without a
task-specific transformation.
\item \textbf{Juxtaposition instead of fusion:} combination tasks place
two sources next to each other rather than inventing a shared structure.
\item \textbf{Surface-level transformation:} the condition decorates an
object but does not change its form, material behavior, or context.
\item \textbf{Verbose but unstable visualization:} the response contains
many adjectives, but readers would not converge on the same image.
\item \textbf{Constraint omission:} adaptation tasks ignore medium,
audience, or domain constraints while producing a plausible generic
design.
\end{itemize}

\begin{methodbox}
\textbf{Appendix use.} These additional results are diagnostic rather
than argumentative. The main paper should remain understandable without
reading this appendix; the appendix supplies implementation details,
replication targets, and failure-analysis scaffolding.
\end{methodbox}

\end{document}